\documentclass{article}

 \usepackage[preprint]{neurips_2026}

\usepackage[utf8]{inputenc} % allow utf-8 input
\usepackage[T1]{fontenc}    % use 8-bit T1 fonts
\usepackage{hyperref}       % hyperlinks
\usepackage{url}            % simple URL typesetting
\usepackage{booktabs}       % professional-quality tables
\usepackage{amsfonts}       % blackboard math symbols
\usepackage{nicefrac}       % compact symbols for 1/2, etc.
\usepackage{microtype}      % microtypography
\usepackage{xcolor}         % colors
\usepackage{amsmath}
\usepackage{graphicx}
\usepackage{subcaption}
\usepackage{colortbl}
\definecolor{lgray}{rgb}{0.85,0.85,0.85}
\definecolor{llgray}{rgb}{0.92,0.92,0.92}
\usepackage{framed}
\usepackage{fvextra}

\renewcommand{\thefootnote}{\fnsymbol{footnote}}
\newcommand{\MethodName}{FAME}

\title{FAME: Forecasting Academic Impact via Continuous-Time Manifold Evolution}

\author{%
  Jianrong Ding, Jianyuan Zhong, Zhengyan Shi, Qiang Xu$^\dagger$ \\
  Department of Computer Science and Engineering \\
  The Chinese University of Hong Kong \\
  \texttt{\{jrding25,jyzhong,zyshi25,qxu\}@cse.cuhk.edu.hk} \\
}

\begin{document}

\maketitle

\begingroup
\renewcommand{\thefootnote}{\fnsymbol{footnote}}
\footnotetext[2]{Corresponding author}
\endgroup

\begin{abstract}
Large Language Models (LLMs) are increasingly used to brainstorm and evaluate research ideas, yet assessing such judgments is fundamentally difficult because the true impact of a new idea may take years to emerge. We address this challenge by using the impact forecasting of human-authored manuscripts as a verifiable proxy task. In a prospective forecasting study, we find that frontier LLMs fail to reliably distinguish high-impact papers from ordinary publications, suggesting that static text-based judging is insufficient for scientific evaluation. To address this limitation, we propose \textbf{\MethodName{}} (\underline{F}orecasting \underline{A}cademic Impact via Continuous-Time \underline{M}anifold \underline{E}volution), a spatiotemporal framework for modeling the dynamic trajectories of scientific topics. \MethodName{} projects papers into a dynamic latent space informed by textual features and a verified knowledge-flow graph, learning geometric constraints that align impactful manuscripts with the forward momentum of their fields. Experiments on 3,200 arXiv papers across three fast-evolving subfields show that \MethodName{} consistently and substantially outperforms state-of-the-art LLM evaluators in prospective multidimensional impact forecasting. Furthermore, integrating \MethodName{}'s dynamic geometric signals into LLMs significantly improves their forecasting performance. These results support manuscript impact forecasting as a useful, measurable proxy benchmark and position \MethodName{} as a strong, trajectory-aware foundation for automated scientific evaluation. The code is available at \url{https://github.com/RafaDD/FAME}.
\end{abstract}
\section{Introduction}

The integration of artificial intelligence into scientific discovery \cite{lu2024ai,yamada2025ai} is accelerating rapidly. Large Language Models (LLMs) are now routinely deployed to navigate literature, brainstorm hypotheses, and assess emerging research directions \cite{lu2024ai,tang2025ai}. As this combinatorial space of possibilities expands, robust automated evaluation becomes essential. Currently, the dominant paradigm relies on LLMs acting as zero-shot or few-shot judges to score novelty and potential impact \cite{lu2024ai,tang2025ai,mishra2025ainstein}. Yet, despite their remarkable fluency in textual critique, a fundamental question remains: \textit{Can LLM evaluators actually distinguish genuinely impactful research from ordinary publications?}

To investigate this, we conducted a rigorous prospective forecasting study on state-of-the-art LLMs \cite{singh2025openaigpt5card,liu2025deepseekv32}. Testing an AI's ability to evaluate novel ideas presents a fundamental methodological challenge: true scientific impact takes much time to materialize, meaning there is no immediate ground truth to verify if the AI's judgment of a brand-new idea is correct. To overcome this, we utilize the impact forecasting of human-authored manuscripts as a verifiable proxy task. By rolling back the clock, treating historical papers as brand-new ideas relative to a temporal cutoff, we can validate an AI's predictions against the papers' actual, realized impact. This establishes impact forecasting as a rigorous proxy benchmark for scientific evaluation, operating on the premise that the ability to forecast the trajectory of a fully formalized manuscript serves as a necessary foundation for effectively assessing raw hypotheses.

Our findings reveal significant limitations in the current baseline. When tasked with predicting the multidimensional impact of published research, frontier LLMs, and even domain-specific models like SciJudge \cite{tong2026ai}, exhibit highly volatile predictive correlations. This suggests that static, text-based judging is insufficient for scientific evaluation. Standard LLMs evaluate text in an isolated semantic vacuum; they overlook the macroscopic spatiotemporal dynamics of a scientific field, missing how a piece of research geometrically aligns with the evolutionary momentum of its underlying topic.

To address this limitation, we argue that impact forecasting must move beyond static text to model the temporal evolution of scientific knowledge. We propose \textbf{\MethodName{}} (\underline{F}orecasting \underline{A}cademic Impact via Continuous-Time \underline{M}anifold \underline{E}volution), a continuous-time spatiotemporal manifold learning framework. \MethodName{} operates in three streamlined stages. First, recognizing that raw citation networks are notoriously noisy and often superficial \cite{cohan2019structural,liang2026citation}, we construct a high-quality inspiration graph using a retrieve-and-verify pipeline, leveraging LLMs strictly to validate genuine foundational knowledge flow between papers. Second, a dual-branch neural architecture maps discrete papers into a continuous manifold while learning dynamic topic spines that track the evolutionary trajectories of sub-disciplines. Finally, we sculpt this latent space using geometric objectives that anchor papers to their temporal context and force high-impact historical manuscripts to align with their field's forward momentum. By structuring the latent space around evolutionary momentum, evaluating an unseen manuscript becomes highly efficient: its predicted impact is derived directly from the geometric alignment between its positional residual and the instantaneous momentum of its corresponding topic spine.
The main contributions of our work are summarized as follows:
\begin{itemize}
    \item \textbf{Empirical Exposure of LLM Limitations:} We systematically quantify the inability of current LLM-as-a-judge paradigms to reliably forecast scientific impact, identifying a critical bottleneck in AI-driven evaluation.
    \item \textbf{Novel Spatiotemporal Framework:} We introduce \textbf{\MethodName{}}, a continuous-time manifold architecture that tracks the evolutionary trajectory of research topics, shifting the evaluation paradigm from static semantic scoring to dynamic geometric alignment.
    \item \textbf{Robust Predictive Performance:} Across extensive sliding-window evaluations on 3,200 arXiv papers, \MethodName{} achieves a Spearman rank correlation exceeding 0.5, consistently and significantly outperforming state-of-the-art LLM baselines. Furthermore, integrating \MethodName{}'s manifold scores into standard LLMs significantly boosts their performance.
\end{itemize}

\begin{figure}[t]
    \centering
    \includegraphics[width=0.95\linewidth]{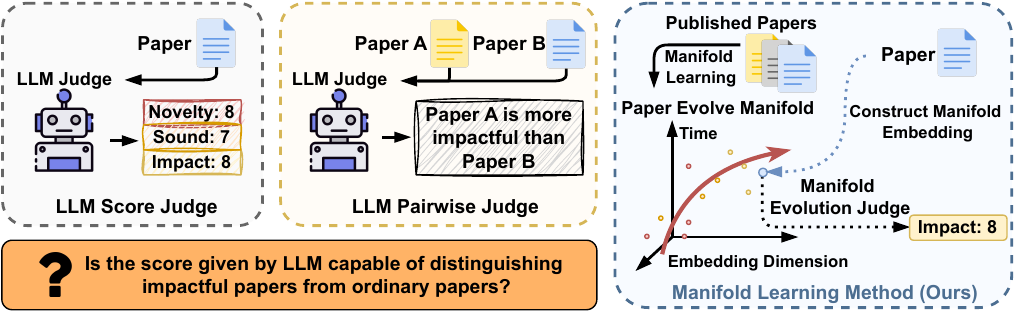}
    \caption{Comparison of standard LLM evaluators with our proposed method, which evaluates research manuscripts based on their geometric alignment with a field's evolutionary momentum.}
    \label{fig:motivation}
\end{figure}
\section{Preliminary}
\label{sec:preliminary}

\textbf{Problem Definition.}
Let $\mathcal{P}=\{p_{1},...,p_{N}\}$ denote a corpus of N scientific papers published over a continuous time horizon. Each paper $p_i$ is associated with a publication timestamp $t_i$ and a realized ground-truth impact weight $w_i$. By computing $w_i$ using bibliometric signals (e.g., citations, GitHub stars) aggregated at the study's global conclusion ($T_{end}$), we frame the task as a calibrated prospective forecast: projecting early manuscript features to their ultimate long-term impact. To simulate a realistic forecasting scenario, we temporally partition the corpus based on a threshold timestamp $T$. Specifically, we define a training set comprising strictly historical publications, $\mathcal{P}_{train}=\{p_{i}\in\mathcal{P}|t_{i}\le T\}$, and a temporally disjoint test set representing future publications, $\mathcal{P}_{test}=\{p_{i}\in\mathcal{P}|t_{i}>T\}$. The overarching objective of this task is to leverage the historical data in $\mathcal{P}_{train}$ to learn a model $\mathcal{M}$ capable of estimating the future realized significance of emerging research. Formally, the model aims to maximize the Spearman rank correlation $\rho_s$ between the predicted impact weights $\hat{\mathbf{w}}_{test}$ and the true impact weights $\mathbf{w}_{test}$, evaluated on the test set:
\begin{equation}
    \max_{\mathcal{M}} \rho_s(\boldsymbol{w}_{test}, \hat{\boldsymbol{w}}_{test}), \quad \text{where} \quad \hat{\boldsymbol{w}}_{test} = \mathcal{M}(\mathcal{P}_{test})
\end{equation}

\textbf{Impact Weight Formulation.}
To construct the ground-truth impact weight $w_i$ of each paper, we aggregate multiple real-world signals that capture different facets of a paper's influence. For each paper $p_i$, let $s_{i,m}$ denote its raw score across $M=4$ distinct metrics: GitHub stars (software utility), citation counts (broad academic attention), influential citations (foundational impact), and Altmetric scores (social and media traction). To account for the heavy-tailed distribution typical of scientific impact metrics \cite{radicchi2008universality,albarran2011skewness}, we apply a scale factor $a_m$ and log-normalization with base \(b_m\) to each individual signal. The final composite impact weight $w_i$ is formulated as the sum of these four normalized scores: $ w_i = \sum_{m=1}^{4} \log_{b_m}(a_m s_{i,m} + 1)$.

\section{Methodology}
\label{sec:methodology}

\begin{figure}[!t]
    \centering
    \includegraphics[width=0.95\textwidth]{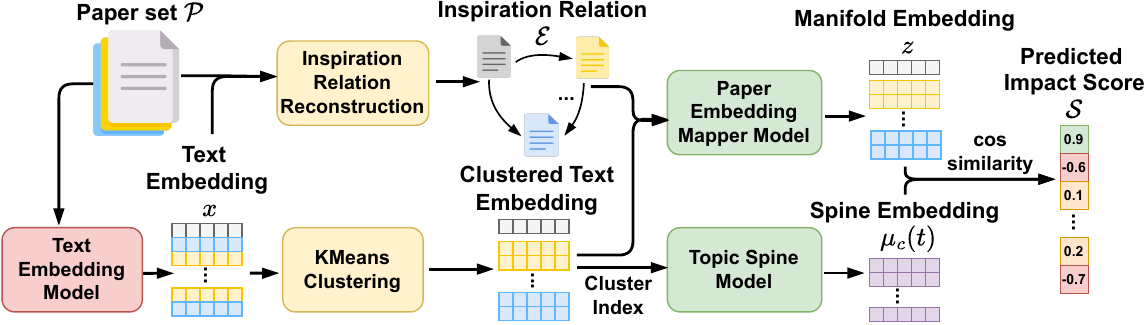}
    \caption{Overall architecture of \MethodName{}. First, we construct a high-quality inspiration graph. Second, a dual-branch neural architecture maps discrete papers into a continuous manifold. Finally, we predict the impact score using manifold embeddings.}
    \label{fig:pipeline}
\end{figure}

To address the problem of scientific impact prediction, we propose \MethodName{}, a continuous-time framework that models the evolution of scientific topics as dynamic trajectories within a latent space. As shown in Figure \ref{fig:pipeline}, the framework operates in three sequential stages: \textbf{(1)} distilling noisy academic citations into a high-quality, verified inspiration graph; \textbf{(2)} mapping raw textual features into a spatiotemporal manifold using continuous representation learning; and \textbf{(3)} leveraging the geometric properties of the learned manifold to infer the future impact of unseen papers.

\subsection{Inspiration Graph Construction}

\textbf{Semantic Clustering.} We first utilize a pre-trained text embedding model to transform the raw textual content of the corpus into dense semantic representations, denoted as $\{x_i\}_{i=1}^N$. To capture the underlying sub-disciplines, we apply KMeans clustering on these embeddings, assigning a base topic label $c_i \in \{1, \dots, K\}$ to each paper. Each cluster represents a localized scientific topic within the broader academic corpus.

\textbf{Retrieve-and-Verify Graph Construction.} Raw citation networks are notoriously noisy; many citations are superficial and do not reflect genuine intellectual inspiration \cite{cohan2019structural,liang2026citation}. To isolate meaningful knowledge transmission, we construct a high-quality inspiration graph $\mathcal{E}$ using a rigorous retrieve-and-verify pipeline. For a given target paper $p_j$, we first retrieve a candidate precursor pool $\{p_i\}$ using three strict heuristic constraints: \textbf{(1) Citation Link:} $p_i$ must be explicitly present in the bibliography of $p_j$. \textbf{(2) Semantic Similarity:} The cosine similarity between the embeddings of the two papers must satisfy a minimum threshold: $\cos(x_i, x_j) > \tau_{sim}$. \textbf{(3) Temporal Precedence:} The publication time difference must enforce strict chronological order: $t_j - t_i > \Delta_{days}$.
After filtering for the top-$K$ semantically similar candidates, we deploy an LLM to explicitly verify whether $p_j$ was fundamentally inspired by the methodology or findings of $p_i$. Only LLM-verified pairs $(i, j)$ are retained in the final directional edge set $\mathcal{E}$. By utilizing the LLM solely to evaluate the direct textual evidence of inspiration between two localized documents, we effectively leverage its established strengths in textual critique to sanitize the noisy bibliographic data. This produces a refined graph capturing foundational, idea-level relationships rather than mere bibliographic references.

\subsection{Spatiotemporal Manifold Architecture}

\begin{figure}[!t]
    \centering
    \includegraphics[width=0.95\textwidth]{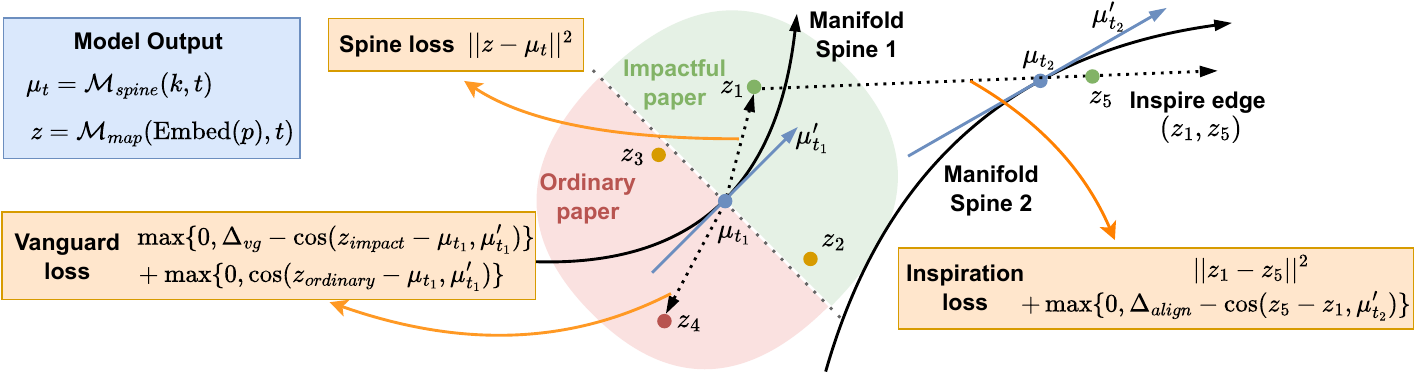}
    \caption{Visualization of the manifold training dynamics and the three applied geometric constraints. The objective functions sculpt the latent space topology by using a spine binding loss to anchor papers to their temporal context, an inspiration loss to ensure knowledge flow alignment, and a vanguard loss to geometrically separate high-impact historical papers from marginal ones.}
    \label{fig:loss}
\end{figure}

As shown in Figure \ref{fig:loss}, we employ a dual-branch neural architecture to map discrete papers into a continuous manifold space and model the overarching temporal development of the topics.

\textbf{Paper Embedding Mapper ($\mathcal{M}_{map}$):} This module parameterized by a Multi-Layer Perceptron (MLP) projects individual papers from the semantic space into the spatiotemporal manifold. It takes the original semantic embedding $x_i$ and the publication timestamp $t_i$ as inputs. To prevent the 1D temporal scalar from being numerically overwhelmed by the high-dimensional text features, we project time into a higher-dimensional space using a continuous temporal encoding function, $\text{TimeEncode}(t_i) \in \mathbb{R}^{d_{time}}$. The features are subsequently fused and projected to yield the final latent spatiotemporal embedding $z_i$:
\begin{equation}
z_i = \mathcal{M}_{map}\left(x_i, \text{TimeEncode}(t_i)\right)
\end{equation}

\textbf{Topic Spine Model ($\mathcal{M}_{spine}$):} This MLP-based module generates the continuous evolutionary trajectory, namely \textit{spine}, for each topic cluster. We assign a learnable, static base representation $E_k \in \mathbb{R}^{d_{topic}}$ to each topic $k \in \{1, \dots, K\}$. The continuous trajectory of the topic is then generated dynamically by passing the base embedding and continuous time $t$ into the model:
\begin{equation}
\mu_k(t) = \mathcal{M}_{spine}\left(E_k, \text{TimeEncode}(t)\right)
\end{equation}

\subsection{Manifold Training Dynamics}

To geometrically structure the latent space, we optimize the topology using three distinct objective functions at the micro and macro levels. The visualization of the losses is shown in Figure \ref{fig:loss}.

\textbf{Spine Binding Loss:} This objective acts as a centripetal structural constraint, anchoring paper embeddings to the evolutionary trajectory of their respective topics at the exact time of publication, thereby preventing arbitrary dispersion in the latent space:
\begin{equation}
\mathcal{L}_{spine} = \sum_{i=1}^{N} \left\| z_i - \mu_{c_i}(t_i) \right\|_2^2
\end{equation}

\textbf{Inspiration Loss:} To model directional knowledge flow for an inspiration relation $(i, j) \in \mathcal{E}$, this loss aligns the geometric knowledge jump $(z_j - z_i)$ with the inspired topic's forward momentum $\mu'_{c_j}(t_j) $, approximated via finite difference as $\mu'_{c_j}(t_j) \approx \mu_{c_j}(t_j) - \mu_{c_j}(t_j - \Delta t)$. The total loss combines an L2 term for spatial proximity (pulling inspiring and inspired papers closer) and a cosine margin term to ensure the innovation direction aligns perfectly with the field's future trajectory.
\begin{equation}
\mathcal{L}_{inspire} = \sum_{(i,j)\in\mathcal{E}} \max\left(0, \Delta_{align} - \cos\left(z_j - z_i, \mu'_{c_j}(t_j)\right)\right) + \sum_{(i,j)\in\mathcal{E}} \left\| z_i - z_j \right\|_2^2
\end{equation}

\textbf{Vanguard Loss:} To integrate ground-truth impact $w_i$, we sculpt the manifold using a localized baseline to isolate genuine breakthroughs from broader temporal drift. For paper $p_i$, we define a spatiotemporal neighborhood $\mathcal{S}_i = \{j \mid c_j = c_i \text{ and } |t_j - t_i| \le \tau_{time}\}$ and compute its median impact $m_i = \text{median}(\{w_j \mid j \in \mathcal{S}_i\})$. Using a dynamic margin $\Delta_{vg}^{(i)} = \Delta_{base} + (1 - \Delta_{base})\tilde{w}_i$, where $\tilde{w}_i$ is min-max normalization of $w_i$ to $[0, 1]$, high-impact vanguard papers ($w_i \ge m_i$) are geometrically encouraged to align with the topic's forward momentum $\mu'_{c_i}(t_i)$. Conversely, low-impact papers ($w_i < m_i$) are penalized if their positional residual projects along this derivative:
\begin{equation}
\begin{aligned}
\mathcal{L}_{vanguard} &= \sum_{i : \tilde{w}_i \ge m_i} \max\left(0, \Delta_{vg}^{(i)} - \cos\left(z_i - \mu_{c_i}(t_i), \mu'_{c_i}(t_i)\right)\right) \\
&\quad + \sum_{i : \tilde{w}_i < m_i} \max\left(0, \cos\left(z_i - \mu_{c_i}(t_i), \mu'_{c_i}(t_i)\right)\right)
\end{aligned}
\end{equation}

The total objective function is a weighted sum of these three losses:
\begin{equation}
   \mathcal{L}_{total} = \alpha\mathcal{L}_{spine} + \beta\mathcal{L}_{inspire} + \gamma\mathcal{L}_{vanguard}
\end{equation}

\subsection{Downstream Impact Inference}

Inference for new papers in \MethodName{} is highly efficient. For an unseen paper $p_{new}$ at time $t_{new}$, we assign its semantic embedding $x_{new}$ to a topic cluster $c_{new}$ via KMeans centroids, and the mapper model projects it to a latent representation $z_{new}$. 
To predict impact, we evaluate the paper's positional residual $(z_{new} - \mu_{c_{new}}(t_{new}))$ against its field's instantaneous momentum $\mu'_{c_{new}}(t_{new})$. Crucially, we quantify this using angular alignment rather than spatial distance. Due to the curse of dimensionality, Euclidean distances converge in high-dimensional spaces \cite{aggarwal2001surprising}, whereas angular metrics remain robust. Thus, the final predicted impact score $\mathcal{S}$ is calculated as their cosine similarity:

\begin{equation}
\mathcal{S}(p_{new}) = \cos\left(z_{new} - \mu_{c_{new}}(t_{new}), \mu'_{c_{new}}(t_{new})\right)
\end{equation}

A score near $1$ indicates strong directional consistency with the field's future evolutionary trajectory, naturally translating to a high predicted impact weight.

\section{Experiments}
\label{sec:experiments}

% \subsection{Experimental Setup}

\begin{figure}[!t]
    \centering
    \includegraphics[width=0.95\linewidth]{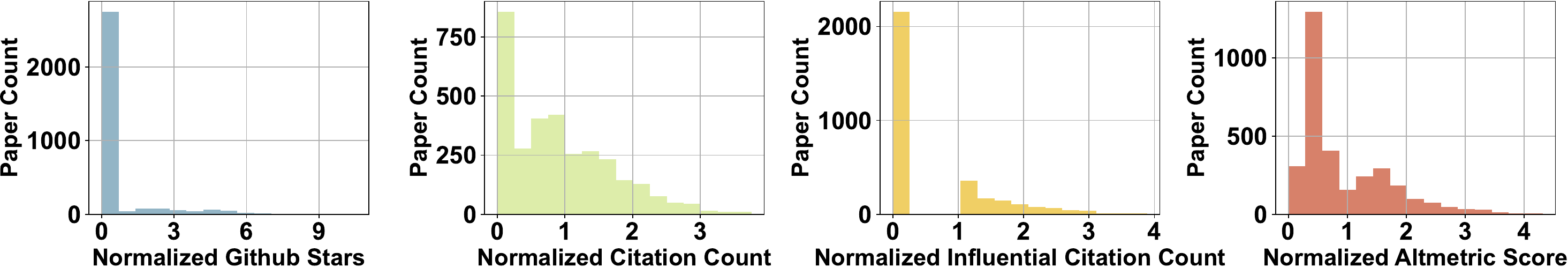}
    \caption{The heavy-tailed distributions of the four raw log-normalized ground-truth metrics used to calculate the composite impact weight for the 3,200 arXiv papers in the dataset.}
    \label{fig:impact}
\end{figure}

\textbf{Dataset and Metrics.} We curate a temporally stamped dataset of 3,200 arXiv papers from three rapidly evolving subfields: Time Series Forecasting, Sparse Attention, and Image Generation. To quantify genuine multi-dimensional influence, we compute a composite ground-truth impact weight $w_i$ for each paper by log-normalizing and aggregating its GitHub stars, citation counts, influential citations, and Altmetric score. The distribution of the four log-normalized ground-truth impact metrics utilized in the dataset is shown in Figure \ref{fig:impact}.

\textbf{Baselines.} We benchmark \MethodName{} against three distinct categories of evaluators. First, we evaluate ML models, including XGBoost \cite{chen2016xgboost}, SVR \cite{cortes1995support,smola2004svr}, Transformer \cite{vaswani2017attention} and TGCN \cite{zhao2019t}, trained directly on raw semantic text embeddings to predict continuous impact weights (we use the inspiration graph as the graph for TGCN). Second, we test general-purpose frontier LLMs (including DeepSeek-V3.2 \cite{liu2025deepseekv32}, GPT-4.1-mini \cite{achiam2023gpt4}, GPT-4.1 \cite{achiam2023gpt4}, GPT-5.2 \cite{singh2025openaigpt5card}, Gemini-3-Flash-Preview, and Gemini-3.1-Pro-Preview), which are evaluated using few-shot calibrated prompting. The prediction of the LLM is an average of predicted novelty, interestingness, and potential impact score, ranging from 1 to 10. Finally, we benchmark against SciJudge-4B \cite{tong2026ai}, a model fine-tuned on Qwen3-4B \cite{yang2025qwen3} for academic comparisons. We adapted its pairwise preference outputs into continuous scores using a tournament-style ranking approach (detailed in Appendix \ref{app:scijudge}).

\textbf{Evaluation Protocol.} 
To emulate prospective forecasting, we evaluate our framework under a temporal out-of-distribution (OOD) setting following SciJudge \cite{tong2026ai}. We employ an 18-month sliding-window evaluation. For each temporal cutoff $T$, models predict the impact of unseen papers published within a 2-month future testing window. While \MethodName{} is trained exclusively on historical data up to $T$, general LLMs suffer from earlier knowledge cutoffs $(T_{LLM}<T)$. To ensure strict parity under our calibrated framework, all evaluators (including \MethodName{}, ML baselines, SciJudge, and few-shot LLMs) are anchored to the exact same realized impact labels. For LLMs, we construct a 30-paper few-shot hint window that strictly pairs historical abstracts ($T_{LLM} < t_{hint} \le T$) with their finalized impact weights collected at $T_{end}$. This shared label access neutralizes temporal disparities, purely isolating each model's capacity to trace early textual indicators to long-term outcomes. Performance is measured via the Spearman rank correlation between predicted and ground-truth impact weights.

\subsection{Main Results}
\begin{table}[!t]
\centering
\caption{Prospective forecasting performance across an 18-month sliding window evaluation from June 2024 to November 2025. Performance is measured by the Spearman rank correlation $\rho_s$ between predicted and ground-truth composite impact weights. The experiments are carried out three times, and the standard deviation of the performance is reported. Best results are shown in bold.}
\label{tab:main}
\resizebox{1\linewidth}{!}{
\begin{tabular}{l|cccccc|c}
\toprule
\toprule 
\textbf{Model} & 2024.06 & 2024.07 & 2024.08 & 2024.09 & 2024.10 & 2024.11 & \textbf{Avg} \\
\midrule
XGBoost & 0.148 $\pm$ 0.027 & 0.321 $\pm$ 0.024 & 0.301 $\pm$ 0.030 & 0.105 $\pm$ 0.035 & 0.182 $\pm$ 0.039 & -0.067 $\pm$ 0.037 & 0.165 \\
SVR & 0.151 $\pm$ 0.005 & 0.225 $\pm$ 0.017 & 0.227 $\pm$ 0.001 & 0.394 $\pm$ 0.004 & 0.281 $\pm$ 0.007 & 0.320 $\pm$ 0.003 & 0.266 \\
Transformer & 0.115 $\pm$ 0.053 & 0.410 $\pm$ 0.037 & 0.397 $\pm$ 0.043 & 0.375 $\pm$ 0.013 & 0.243 $\pm$ 0.036 & 0.127 $\pm$ 0.027 & 0.278 \\
TGCN & 0.351 $\pm$ 0.055 & 0.383 $\pm$ 0.081 & 0.102 $\pm$ 0.063 & 0.207 $\pm$ 0.063 & 0.295 $\pm$ 0.044 & 0.220 $\pm$ 0.084 & 0.260 \\
\midrule
GPT-4.1-mini & 0.237 $\pm$ 0.010 & 0.069 $\pm$ 0.052 & 0.209 $\pm$ 0.033 & 0.269 $\pm$ 0.035 & 0.314 $\pm$ 0.028 & 0.131 $\pm$ 0.103 & 0.205 \\
GPT-4.1 & 0.070 $\pm$ 0.043 & 0.078 $\pm$ 0.029 & 0.076 $\pm$ 0.066 & 0.301 $\pm$ 0.048 & 0.220 $\pm$ 0.033 & 0.021 $\pm$ 0.047 & 0.128 \\
\midrule
\MethodName{} (Ours) & \cellcolor{lgray}{\textbf{0.521 $\pm$ 0.051}} & \cellcolor{lgray}{\textbf{0.633 $\pm$ 0.016}} & \cellcolor{lgray}{\textbf{0.483 $\pm$ 0.006}} & \cellcolor{lgray}{\textbf{0.531 $\pm$ 0.007}} & \cellcolor{lgray}{\textbf{0.562 $\pm$ 0.026}} & \cellcolor{lgray}{\textbf{0.482 $\pm$ 0.036}} & \cellcolor{lgray}{\textbf{0.535}} \\
\midrule
\midrule
\textbf{Model} & 2024.12 & 2025.01 & 2025.02 & 2025.03 & 2025.04 & 2025.05 & \textbf{Avg} \\
\midrule
XGBoost & 0.154 $\pm$ 0.074 & 0.183 $\pm$ 0.056 & 0.298 $\pm$ 0.046 & 0.364 $\pm$ 0.046 & 0.219 $\pm$ 0.047 & 0.362 $\pm$ 0.014 & 0.263 \\
SVR & 0.370 $\pm$ 0.016 & 0.319 $\pm$ 0.024 & 0.467 $\pm$ 0.001 & 0.447 $\pm$ 0.001 & 0.309 $\pm$ 0.003 & 0.479 $\pm$ 0.004 & 0.399 \\
Transformer & 0.257 $\pm$ 0.041 & 0.377 $\pm$ 0.068 & 0.422 $\pm$ 0.035 & 0.062 $\pm$ 0.054 & 0.297 $\pm$ 0.027 & 0.512 $\pm$ 0.018 & 0.321 \\
TGCN & 0.277 $\pm$ 0.042 & 0.248 $\pm$ 0.029 & 0.344 $\pm$ 0.022 & 0.151 $\pm$ 0.053 & 0.212 $\pm$ 0.085 & 0.268 $\pm$ 0.056 & 0.250 \\
\midrule
GPT-4.1-mini & 0.260 $\pm$ 0.030 & 0.048 $\pm$ 0.027 & 0.160 $\pm$ 0.045 & 0.141 $\pm$ 0.039 & 0.051 $\pm$ 0.049 & 0.096 $\pm$ 0.024 & 0.126 \\
GPT-4.1 & 0.106 $\pm$ 0.056 & -0.004 $\pm$ 0.041 & 0.103 $\pm$ 0.062 & 0.058 $\pm$ 0.037 & 0.116 $\pm$ 0.023 & 0.093 $\pm$ 0.035 & 0.079 \\
Gemini-3-Flash-Preview & - & 0.061 $\pm$ 0.027 & 0.149 $\pm$ 0.033 & 0.209 $\pm$ 0.025 & 0.303 $\pm$ 0.022 & 0.259 $\pm$ 0.036 & 0.196 \\
Gemini-3.1-Pro-Preview & - & 0.072 $\pm$ 0.053 & 0.146 $\pm$ 0.016 & 0.243 $\pm$ 0.035 & 0.315 $\pm$ 0.015 & 0.298 $\pm$ 0.018 & 0.215 \\
DeepSeek-V3.2 & - & 0.043 $\pm$ 0.063 & 0.230 $\pm$ 0.050 & 0.255 $\pm$ 0.015 & 0.306 $\pm$ 0.000 & 0.214 $\pm$ 0.060 & 0.210 \\
SciJudge-4B & - & 0.122 $\pm$ 0.006 & 0.156 $\pm$ 0.028 & 0.269 $\pm$ 0.005 & 0.254 $\pm$ 0.004 & 0.312 $\pm$ 0.074 & 0.247 \\
\midrule
\MethodName{} (Ours) & \cellcolor{lgray}{\textbf{0.524 $\pm$ 0.010}} & \cellcolor{lgray}{\textbf{0.433 $\pm$ 0.055}} & \cellcolor{lgray}{\textbf{0.563 $\pm$ 0.025}} & \cellcolor{lgray}{\textbf{0.492 $\pm$ 0.023}} & \cellcolor{lgray}{\textbf{0.450 $\pm$ 0.005}} & \cellcolor{lgray}{\textbf{0.617 $\pm$ 0.029}} & \cellcolor{lgray}{\textbf{0.513}} \\
\midrule
\midrule
\textbf{Model} & 2025.06 & 2025.07 & 2025.08 & 2025.09 & 2025.10 & 2025.11 & \multicolumn{1}{c}{\textbf{Avg}} \\
\midrule
XGBoost & 0.347 $\pm$ 0.066 & 0.037 $\pm$ 0.068 & 0.288 $\pm$ 0.038 & 0.169 $\pm$ 0.024 & 0.061 $\pm$ 0.049 & 0.054 $\pm$ 0.047 & 0.160 \\
SVR & 0.459 $\pm$ 0.003 & 0.200 $\pm$ 0.000 & 0.270 $\pm$ 0.004 & 0.320 $\pm$ 0.000 & 0.297 $\pm$ 0.001 & 0.238 $\pm$ 0.015 & 0.297 \\
Transformer & 0.507 $\pm$ 0.028 & 0.441 $\pm$ 0.044 & 0.454 $\pm$ 0.055 & 0.350 $\pm$ 0.051 & 0.057 $\pm$ 0.061 & -0.047 $\pm$ 0.015 & 0.294 \\
TGCN & 0.077 $\pm$ 0.041 & 0.409 $\pm$ 0.071 & 0.508 $\pm$ 0.033 & 0.534 $\pm$ 0.019 & 0.355 $\pm$ 0.031 & -0.155 $\pm$ 0.026 & 0.288 \\
\midrule
GPT-4.1-mini & 0.009 $\pm$ 0.016 & 0.151 $\pm$ 0.029 & 0.174 $\pm$ 0.060 & 0.140 $\pm$ 0.047 & 0.251 $\pm$ 0.146 & 0.143 $\pm$ 0.028 & 0.145 \\
GPT-4.1 & -0.131 $\pm$ 0.064 & -0.057 $\pm$ 0.017 & -0.184 $\pm$ 0.070 & 0.103 $\pm$ 0.034 & 0.220 $\pm$ 0.090 & 0.070 $\pm$ 0.036 & 0.003 \\
Gemini-3-Flash-Preview & 0.142 $\pm$ 0.050 & 0.263 $\pm$ 0.081 & 0.060 $\pm$ 0.092 & 0.197 $\pm$ 0.023 & 0.216 $\pm$ 0.041 & 0.231 $\pm$ 0.058 & 0.185 \\
Gemini-3.1-Pro-Preview & 0.005 $\pm$ 0.034 & 0.088 $\pm$ 0.072 & 0.104 $\pm$ 0.038 & 0.280 $\pm$ 0.044 & 0.140 $\pm$ 0.024 & 0.168 $\pm$ 0.037 & 0.131 \\
DeepSeek-V3.2 & 0.156 $\pm$ 0.030 & 0.345 $\pm$ 0.084 & 0.134 $\pm$ 0.037 & 0.237 $\pm$ 0.085 & 0.129 $\pm$ 0.078 & 0.143 $\pm$ 0.112 & 0.191 \\
GPT-5.2 & - & - & 0.082 $\pm$ 0.031 & 0.250 $\pm$ 0.033 & 0.200 $\pm$ 0.024 & 0.183 $\pm$ 0.048 & 0.179 \\
SciJudge-4B & 0.174 $\pm$ 0.112 & 0.419 $\pm$ 0.012 & 0.226 $\pm$ 0.055 & 0.213 $\pm$ 0.025 & 0.274 $\pm$ 0.069 & 0.178 $\pm$ 0.136 & 0.302 \\
\midrule
\MethodName{} (Ours) & \cellcolor{lgray}{\textbf{0.616 $\pm$ 0.026}} & \cellcolor{lgray}{\textbf{0.676 $\pm$ 0.027}} & \cellcolor{lgray}{\textbf{0.582 $\pm$ 0.013}} & \cellcolor{lgray}{\textbf{0.552 $\pm$ 0.032}} & \cellcolor{lgray}{\textbf{0.399 $\pm$ 0.023}} & \cellcolor{lgray}{\textbf{0.370 $\pm$ 0.052}} & \cellcolor{lgray}{\textbf{0.532}} \\
\bottomrule
\bottomrule
\end{tabular}}
\end{table}

As detailed in our comparative evaluations across the 18-month horizon shown in Table \ref{tab:main}, \MethodName{} significantly and consistently outperforms all baseline models. By mapping papers to continuous spatiotemporal trajectories, \MethodName{} achieves average Spearman correlations exceeding 0.50. 
In contrast, machine learning baselines (XGBoost, SVR, Transformer, and TGCN) show moderate but limited performance. Because they rely entirely on static text embeddings, they fail to capture the temporal momentum of the field. Standard LLMs exhibit poorer and more volatile performance ($\rho_s < 0.25$), frequently fluctuating near random correlation. This highlights the severe limitations of using static, text-based LLM evaluators for prospective impact prediction. While the domain-specific SciJudge-4B demonstrates improved stability over general LLMs, its reliance on pairwise text comparisons lacks macro-level trajectory awareness, leaving it substantially short of \MethodName{}'s performance. We also examine the top-5 accuracy of the predicted papers; the results are shown in Appendix \ref{app:topk_acc}. Furthermore, we demonstrate that integrating our continuous-time manifold scores as explicit context into standard LLM evaluators significantly enhances their predictive performance, as detailed in Appendix \ref{app:manifold_integration}.

\subsection{Ablation Study}

To rigorously validate the contribution of each component within our model architecture, we conduct quantitative, qualitative ablation studies and sensitivity analysis of loss weights.

\textbf{Quantitative Analysis.} We compare the full \MethodName{} against three variants: removing the inspiration loss ($\mathcal{L}_{inspire}$), the vanguard loss ($\mathcal{L}_{vanguard}$), and the spine loss ($\mathcal{L}_{spine}$). 
We also test a naive method which trains the scoring model directly on the original text embeddings of the paper produced by a text embedding model. To validate our retrieve-and-verify pipeline, we replace the verified inspiration graph with two baselines: a raw, unfiltered citation graph and a random inspiration graph using eight prior publications as semantic inspirations. The removal of any structural objective leads to a pronounced degradation in predictive performance. Most notably, dropping the vanguard loss yields the most severe performance drop. This confirms that localized impact normalization is critical. It indicates that, without explicitly encouraging high-impact papers to align with the forward momentum of the field, the model cannot effectively rank future contributions. The detailed results are shown in Appendix \ref{app:ablation_on_loss} and Table \ref{tab:ablation}.

\begin{figure}[!t]
    \centering
    
    \begin{subfigure}{0.48\textwidth}
        \centering
        \includegraphics[width=\linewidth]{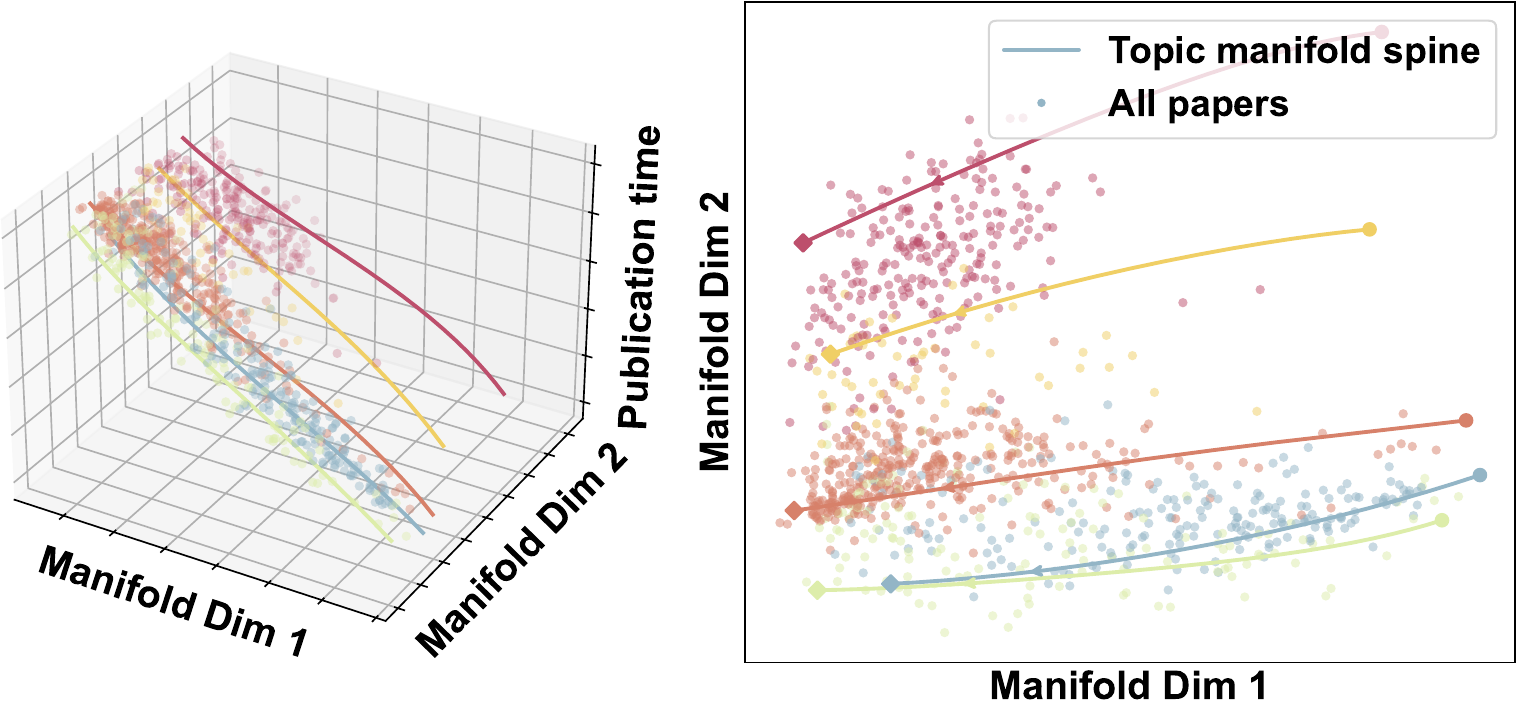}
        \caption{Manifold trained with all losses.}
        \label{fig:sub1}
    \end{subfigure}
    \hfill
    \begin{subfigure}{0.48\textwidth}
        \centering
        \includegraphics[width=\linewidth]{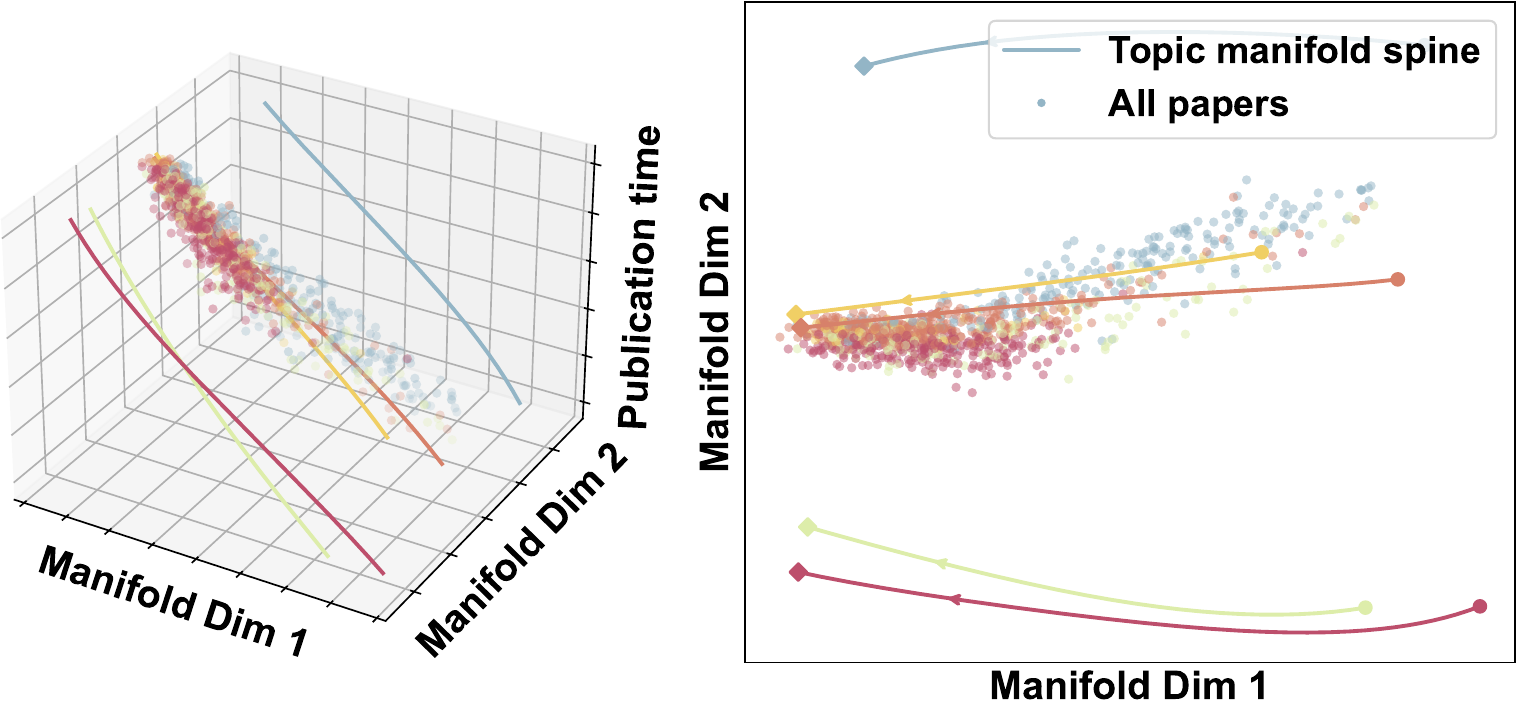}
        \caption{Manifold trained without spine loss.}
        \label{fig:sub2}
    \end{subfigure}
    
    \vspace{0.5cm}
    
    \begin{subfigure}{0.48\textwidth}
        \centering
        \includegraphics[width=\linewidth]{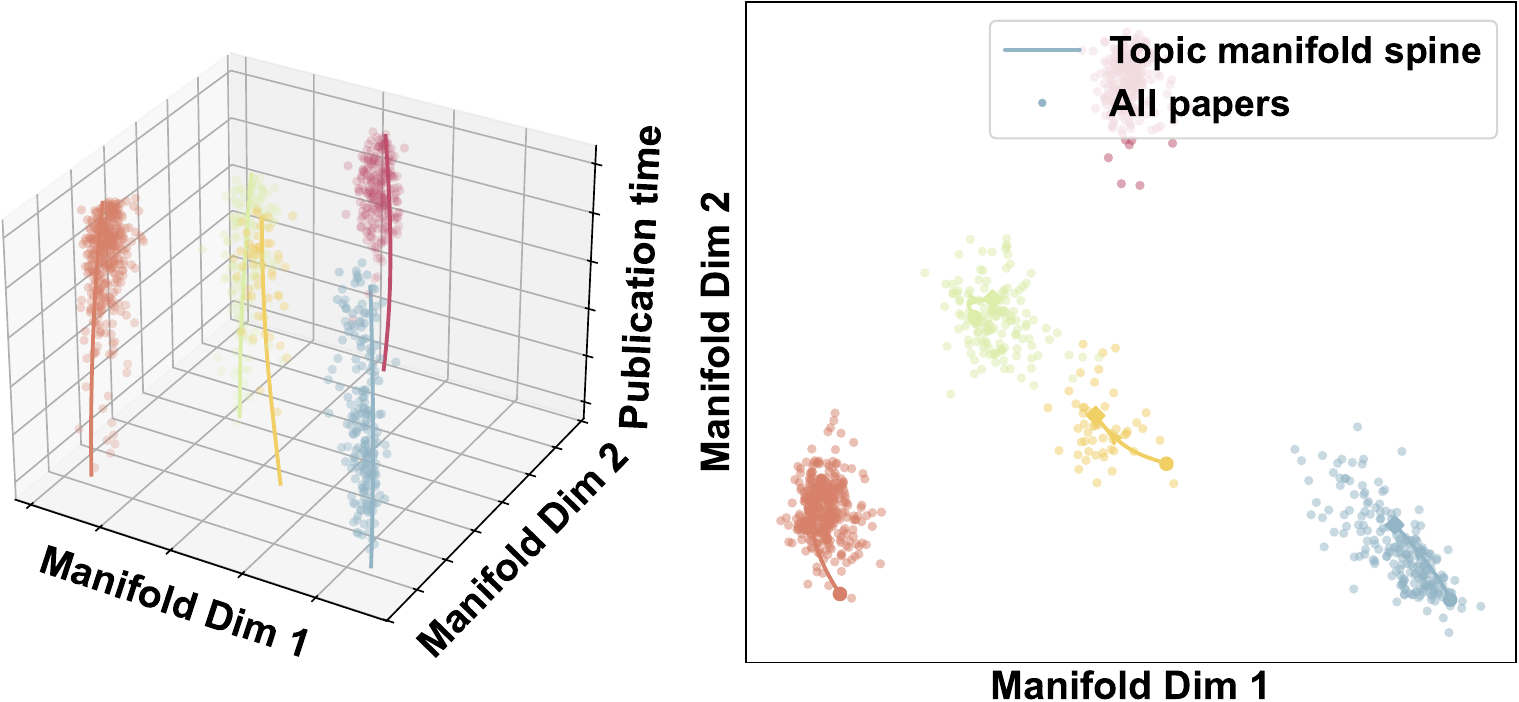}
        \caption{Manifold trained without inspiration loss.}
        \label{fig:sub3}
    \end{subfigure}
    \hfill
    \begin{subfigure}{0.48\textwidth}
        \centering
        \includegraphics[width=\linewidth]{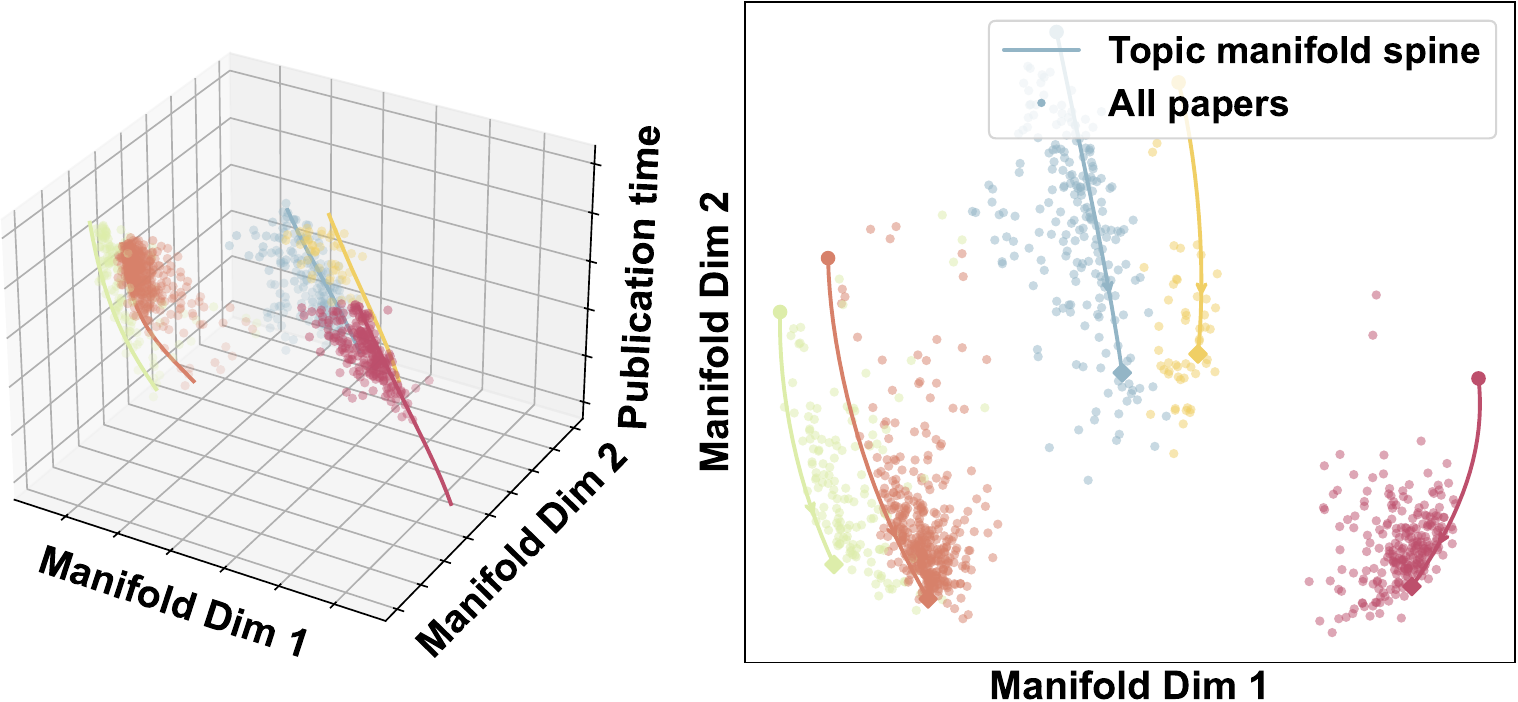}
        \caption{Manifold trained without vanguard loss.}
        \label{fig:sub4}
    \end{subfigure}
    
    \caption{Qualitative ablation of the spatiotemporal latent manifold. The panels demonstrate how removing individual geometric constraints disrupts the structural integrity of the field's evolutionary trajectory. Within each panel, the left plot illustrates the 3D evolution of the manifold along the publication time axis, while the right plot displays a 2D projection of the embeddings with the temporal dimension collapsed. Different clusters of papers are shown in different colors.}
    \label{fig:ablation}
\end{figure}

\textbf{Qualitative Latent Space Dynamics.} We used KernelPCA \cite{scholkopf1997kernelpca} to reduce the dimension of the embeddings. As shown in Figure \ref{fig:ablation}, the fully optimized manifold produces continuous, well-aligned topic trajectories. Removing the spine loss destroys the global temporal anchor. Removing the inspiration loss collapses papers into static clusters devoid of directional knowledge flow. Finally, removing the vanguard loss results in a uniform spatial distribution that utterly fails to differentiate breakthrough research from marginal publications.

\textbf{Sensitivity Analysis.} To evaluate the interplay between \MethodName{}'s geometric constraints, we conducted a sensitivity analysis on the loss formulation weights (detailed in Appendix \ref{app:loss_w_sensitivity}). By systematically varying the spine binding ($\alpha$), inspiration ($\beta$), and vanguard ($\gamma$) loss weights, we found that the framework achieves strong, stable predictive performance when $\beta$ is relatively small, provided $\alpha$ and $\gamma$ are located within a reasonable range. This confirms that while all three structural objectives are necessary, balancing macroscopic temporal anchoring and localized impact normalization is the primary driver of the model's forecasting efficacy.
\section{Discussion}
\label{sec:discussion}

\textbf{Rationality of proposed impact score.} Our ground-truth impact weight aggregates GitHub stars, citations, influential citations, and Altmetric scores to capture multi-dimensional scientific influence. To validate this metric, we randomly sampled 300 papers and obtained ground-truth ratings from three independent human evaluators based on novelty, soundness, and potential impact. Comparing our automated metrics to the average human rating, Table \ref{tab:human_score_relation} shows that the aggregated Impact Score correlates more strongly with human consensus than any individual metric alone. Furthermore, Appendix \ref{app:corr_human} illustrates that this strong alignment holds consistently across subfields. This confirms our composite score serves as a robust, human-aligned proxy for genuine scientific impact.

\begin{table}[!t]
\centering
\caption{Pearson and Spearman correlation evaluating the alignment between the proposed impact metrics and human ratings of the sampled 300 papers. The results demonstrate that the combined Impact Score aligns more strongly with human consensus than any individual normalized component.}
\label{tab:human_score_relation}
\resizebox{0.8\linewidth}{!}{
\begin{tabular}{l|cc}
\toprule
\toprule 
 \textbf{Normalized Score Component} &  \textbf{Pearson Correlation $\rho$} &  \textbf{Spearman Correlation $\rho_s$} \\
 \midrule
Github Stars & 0.4032 & 0.3020 \\
Citation Count & 0.5277 & 0.4289 \\
Influential Citation Count & 0.5673 & 0.4884 \\
Altmetric Score & 0.4677 & 0.4169 \\
\midrule
Impact score (Sum four scores) & \cellcolor{lgray}{\textbf{0.6549}} & \cellcolor{lgray}{\textbf{0.5283}} \\
\bottomrule
\bottomrule
\end{tabular}}
\end{table}

\begin{figure}[!t]
    \centering
    \includegraphics[width=\linewidth]{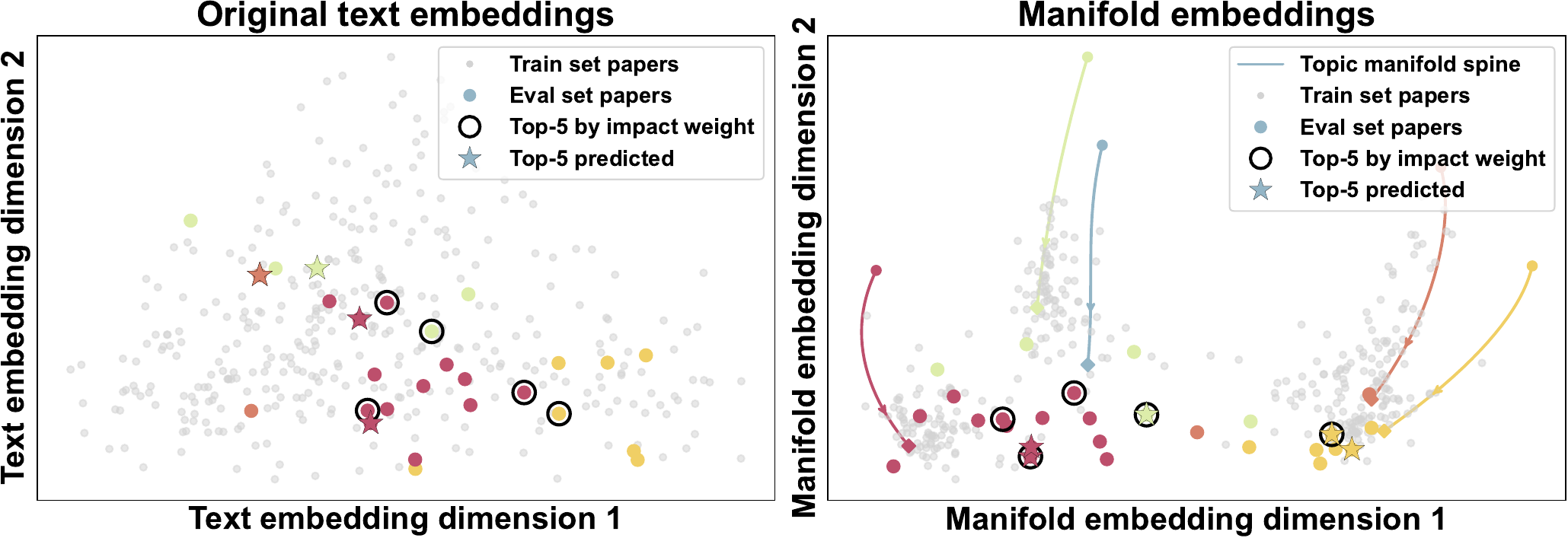}
    \caption{2D visualizations comparing the latent spaces of standard text embeddings versus the \MethodName{} manifold. The left panel illustrates a geometric bottleneck in raw text embeddings, where novel breakthrough ideas are located alongside ordinary publications. The right panel demonstrates how \MethodName{} resolves this problem, yielding top-5 impactful predictions that align significantly better with the true ground-truth impact weights.}
    \label{fig:raw_vs_manifold_0}
\end{figure}

\textbf{Comparison between raw text embeddings and \MethodName{}.} Our prospective forecasting results expose a critical flaw in static text evaluations: they ignore a field's macroscopic evolutionary trajectory. To visualize this limitation, we use KernelPCA to reduce the dimensionality of the embeddings. As illustrated in Figure \ref{fig:raw_vs_manifold_0} (left), standard text embedding models suffer from a geometric bottleneck, collapsing new paper evaluations directly into the dense semantic centroids of historical data. In this compacted space, novel, high-impact research is severely entangled with ordinary work, causing the baseline model's top predictions (Top-5 predicted, denoted by stars) to land seemingly at random and fail to align with actual breakthrough research (true Top-5 papers, denoted by circles). However, our continuous-time spatiotemporal manifold resolves this centroid collapse, as shown in Figure \ref{fig:raw_vs_manifold_0} (right). By projecting textual features into a dynamic latent space structured by continuous topic spines, papers are distributed structurally along their field's evolutionary trajectory. This shifts the evaluation paradigm from static semantic similarity to dynamic geometric alignment, preventing breakthrough ideas from being buried in past literature. As a result, our manifold embeddings allow the predicted vanguard papers (stars) to align closer to the true high-impact ground-truth papers (circles). We provide quantitative analysis in Appendix \ref{app:topk_acc} and additional visual examples in Appendix \ref{app:text_vs_manifold}.

\section{Limitations}
\label{sec:limitations}

While \MethodName{} provides a robust framework for automated scientific evaluation, several limitations remain. First, evaluating formalized arXiv manuscripts as proxies leaves a gap in assessing the nascent, early-stage hypotheses required for end-to-end autonomous AI pipelines. Second, while effective in specific computer science subfields, the framework requires adaptation for disciplines with vastly different publication dynamics and time-to-impact horizons. Methodologically, using finalized impact metrics for historical training data deviates from strict real-time forecasting; future work should explore dynamically evolving, time-truncated target labels at the cutoff $T$. Finally, because citation signals are imperfect proxies for immediate quality, future iterations must model long-term dynamics to accurately capture the delayed impact trajectories of highly innovative research, often termed sleeping beauties.
\section{Related Works}

\textbf{LLM-Driven Scientific Discovery and the Evaluation Bottleneck.} Automated scientific discovery is shifting from domain-specific models \cite{boiko2023autonomous,bran2023chemcrow} to generalist AI Scientist agents driven by iterative generation-evaluation loops \cite{lu2024ai,tang2025ai,yamada2025ai,baek2025researchagent,gottweis2025towards}. While LLMs effectively brainstorm \cite{si2024can}, navigate literature \cite{lala2023paperqa,taylor2022galactica}, and write code \cite{wu2024autogen,yang2023large}, their evaluation modules—reliant on the LLM-as-a-judge paradigm \cite{zheng2023judging,bubeck2023sparksartificialgeneralintelligence}—remain a critical bottleneck. Despite their fluency, LLMs act as myopic evaluators \cite{liu2024lost,wang2024large}; even models fine-tuned for academic comparisons, like SciJudge \cite{tong2026ai}, fail to reliably forecast impact because they ignore the macroscopic evolutionary trajectories of research fields.

\textbf{Spatiotemporal Manifold Learning.} Recent advances in representation learning emphasize Continuous-Time Dynamic Graphs \cite{rossi2020temporal,yu2023towards} and spatiotemporal manifolds \cite{chamberlain2021beltrami,li2023hybrid}, but these architectures are primarily optimized for microscopic node interactions \cite{qu2020continuous}. When applied to scientific impact forecasting—which traditionally relies on static bibliometrics \cite{wang2013quantifying} or retrospective Graph Neural Networks \cite{han2021temporal}—they treat network evolution as a passive aggregation of local events, missing the broader macroscopic trajectory of the domain. By explicitly aligning high-impact papers with their field's momentum, our approach streamlines idea evaluation into a direct geometric similarity calculation between a paper's positional residual and its topic's instantaneous trajectory.

\section{Conclusion}
\label{sec:conclusion}

LLMs are increasingly used to evaluate novel research ideas, but our study reveals they fail to reliably distinguish truly impactful research from ordinary publications. This vulnerability stems from their tendency to evaluate ideas in a static semantic vacuum, overlooking the macroscopic evolutionary trajectory of a scientific field. To address this, we introduce \MethodName{}, a continuous-time spatiotemporal manifold learning framework. By projecting papers into a dynamic latent space structured by continuous topic spines, \MethodName{} evaluates ideas based on their geometric alignment with a field's evolutionary momentum. Extensive sliding-window evaluations on 3,200 arXiv papers demonstrate that \MethodName{} significantly outperforms state-of-the-art LLM evaluators. Furthermore, integrating our manifold scores directly into standard LLMs as explicit context substantially boosts their predictive reasoning. Ultimately, our study establishes manuscript impact forecasting as a rigorous proxy benchmark, positioning \MethodName{} as a robust foundation for scientific evaluation.

{
\bibliographystyle{plain}
\bibliography{bibitems}
}
\clearpage
\appendix

\section{Model Details}
\label{app:model_details}

\subsection{Hyperparameters and Computation Resources}

The hyperparameters for constructing the inspiration graph, training the manifold architecture, and constructing the ground truth impact metrics are detailed in Table \ref{tab:hyperparameters}. These parameters govern both the strict retrieval of candidate papers and the subsequent temporal geometric embeddings.

\begin{table}[htbp]
    \centering
    \caption{Hyperparameter settings for the \MethodName{} manifold architecture and inspiration graph.}
    \begin{tabular}{ll}
        \toprule
        \toprule
        % \textbf{Hyperparameter} & \textbf{Value} \\
        % \midrule
        \multicolumn{2}{c}{\textit{Inspiration Graph (Retrieve-and-Verify)}} \\
        \midrule
        Top-K Candidates per Paper & 12 \\
        Minimum Time Delta ($\Delta_{days}$) & 60 days \\
        Semantic Similarity Threshold ($\tau_{sim}$) & 0.6 \\
        LLM Confidence Threshold & 0.6 \\
        Text Embedding Model & OpenAI text-embedding-3-large \\
        Inspiration Graph Judge LLM & gpt-4.1-mini \\
        \midrule
        \midrule
        \multicolumn{2}{c}{\textit{Manifold Training Defaults}} \\
        \midrule
        Learning Rate & 0.0001 \\
        Finite Difference Step ($\Delta t$) & 0.01 \\
        Inspiration Alignment Similarity ($\Delta_{align}$) & 0.5 \\
        Base Vanguard Similarity ($\Delta_{base}$) & 0 \\
        Temporal Encoding Dimension ($d_{time}$) & 64 \\
        Topic Spine Dimension ($d_{topic}$) & 128 \\
        Spine Loss Weight ($\alpha$) & 1.0 \\
        Inspiration Loss Weight ($\beta$) & 0.1 \\
        Vanguard Loss Weight ($\gamma$) & 1.0 \\
        \midrule
        \midrule
        \multicolumn{2}{c}{\textit{Metrics Construction parameters}} \\
        \midrule
        Github Stars ($a_1, b_1$) & (1, $\mathrm{e}$) \\
        Citation Counts ($a_2, b_2$) & (1, 8)  \\
        Influential Citation Counts ($a_3, b_3$) & (10, 8) \\
        Altmetric Score ($a_4, b_4$) & (1, 4) \\
        \bottomrule
        \bottomrule
    \end{tabular}
    \label{tab:hyperparameters}
\end{table}

All the experiments are conducted on an Ubuntu 22.04 server equipped with an AMD EPYC 7K62 CPU and an NVIDIA A100-64G GPU. The running time for each manifold training process is around 2 minutes.

\subsection{SciJudge Evaluation Method}
\label{app:scijudge}

Because SciJudge-4B \cite{tong2026ai} outputs pairwise preferences rather than continuous scores, we adapt it using a tournament-style ranking approach: for a testing window of $n$ papers, we query the model with all $n(n-1)/2$ possible pairs to assess their relative contributions. We maintain a global winning table, and the aggregate win count for each paper serves as its final predicted impact score.

\subsection{LLM Prompt Templates}

% \vspace{1em}
\noindent\textbf{Prompt Template 1: Inspiration Graph Verification. }
This prompt is used by the judge model gpt-4.1-mini to establish genuine knowledge-flow edges between chronologically ordered paper pairs based on the candidates filtered.

\begin{leftbar}
\begin{Verbatim}[
  breaklines=true,
  breaksymbolleft={},   % remove symbol at start of wrapped line
  breaksymbolright={}   % (optional) remove symbol at end too
]
You judge whether a later paper is inspired by an earlier paper.
Return strict JSON:
{
  "is_inspired": true,
  "confidence": 0.0,
  "rationale": "1-2 sentence explanation"
}

Later paper A:
- arXiv ID: {later.paper.arxiv_id}
- Title: {later.paper.title}
- Abstract: {later.paper.abstract}

Earlier paper B:
- arXiv ID: {earlier.paper.arxiv_id}
- Title: {earlier.paper.title}
- Abstract: {earlier.paper.abstract}

Additional signal:
- cosine_similarity: {similarity}
- time_delta_days(A_minus_B): {time_delta_days}
\end{Verbatim}
\end{leftbar}

% \vspace{1em}
\noindent\textbf{Prompt Template 2: Baseline Impact Forecasting. }This prompt queries the baseline LLM evaluators to score future papers across metrics.

\begin{leftbar}
\begin{Verbatim}[
  breaklines=true,
  breaksymbolleft={},   % remove symbol at start of wrapped line
  breaksymbolright={}   % (optional) remove symbol at end too
]
You are a strict research idea evaluator.
Topic: {topic}

Reference: Use the following papers published after your knowledge cutoff as reference to calibrate your scoring of novelty and potential impact:
{hint_paper.title}{hint_paper.abstract}, impact weight: {hint_paper.impact_weight}
...

For each idea, provide:

- novelty_score (1-10): Originality relative to existing literature; is it non-obvious and new?
- interestingness_score (1-10): Intellectual appeal and whether it raises compelling questions 
  or insights
- potential_impact_score (1-10): Expected influence on the field if successful (e.g., advancing 
  theory, shifting paradigms, enabling new research directions, or widely adopted applications)

Scoring Guidelines:

- Use only integers from 1 (very low) to 10 (exceptional)
- Be conservative; avoid inflated scores
- Compare against typical research in the field, not absolute ideals

Return concise evaluation_notes (1-3 sentences) per idea.
Output strict JSON only:
{
  "evaluations": [
    {
      "index": 0,
      "novelty_score": 7,
      "interestingness_score": 8,
      "potential_impact_score": 7,
      "evaluation_notes": "..."
    }
  ]
}

Ideas to evaluate:
{papers}
\end{Verbatim}
\end{leftbar}

\section{Top-5 Accuracy Evaluation}
\label{app:topk_acc}

We provide a detailed evaluation of the models based on their top-5 prediction accuracy. Specifically, Table \ref{tab:top} illustrates the prospective forecasting performance over an 18-month sliding window from June 2024 to November 2025. This metric measures how accurately a model's top-5 predicted high-impact papers align with the actual ground-truth top-5 papers based on their composite impact weights. The results clearly demonstrate that the proposed \MethodName{} framework consistently outperforms all baseline evaluators, including traditional machine learning models and frontier LLMs across every temporal window.

\begin{table}[ht]
\centering
\caption{Prospective forecasting performance across an 18-month sliding window evaluation from June 2024 to November 2025. Performance is measured by the top-5 accuracy between predicted and ground-truth composite impact weights. The experiments are carried out three times, and the best results are shown in bold.}
\label{tab:top}
\resizebox{1\linewidth}{!}{
\begin{tabular}{l|cccccc|c}
\toprule
\toprule 
\textbf{Model} & 2024.06 & 2024.07 & 2024.08 & 2024.09 & 2024.10 & 2024.11 & \textbf{Avg} \\
\midrule
XGBoost & 0.067 $\pm$ 0.094 & 0.200 $\pm$ 0.000 & 0.000 $\pm$ 0.000 & 0.000 $\pm$ 0.000 & 0.000 $\pm$ 0.000 & 0.067 $\pm$ 0.094 & 0.056 \\
SVR & 0.133 $\pm$ 0.094 & 0.400 $\pm$ 0.000 & 0.400 $\pm$ 0.000 & 0.400 $\pm$ 0.000 & 0.067 $\pm$ 0.094 & 0.133 $\pm$ 0.094 & 0.256 \\
Transformer & 0.200 $\pm$ 0.000 & 0.400 $\pm$ 0.000 & 0.133 $\pm$ 0.094 & 0.200 $\pm$ 0.000 & 0.200 $\pm$ 0.000 & 0.200 $\pm$ 0.000 & 0.222 \\
TGCN & 0.200 $\pm$ 0.000 & 0.067 $\pm$ 0.094 & 0.133 $\pm$ 0.094 & 0.133 $\pm$ 0.094 & 0.200 $\pm$ 0.000 & 0.133 $\pm$ 0.094 & 0.144 \\
\midrule
GPT-4.1-mini & 0.267 $\pm$ 0.094 & 0.200 $\pm$ 0.000 & 0.133 $\pm$ 0.094 & 0.200 $\pm$ 0.000 & 0.200 $\pm$ 0.000 & 0.267 $\pm$ 0.094 & 0.211 \\
GPT-4.1 & 0.067 $\pm$ 0.094 & 0.133 $\pm$ 0.094 & 0.133 $\pm$ 0.094 & 0.133 $\pm$ 0.094 & 0.200 $\pm$ 0.000 & 0.200 $\pm$ 0.000 & 0.144 \\
\midrule
\MethodName{} (Ours) & \cellcolor{lgray}{\textbf{0.600 $\pm$ 0.000}} & \cellcolor{lgray}{\textbf{0.800 $\pm$ 0.000}} & \cellcolor{lgray}{\textbf{0.733 $\pm$ 0.094}} & \cellcolor{lgray}{\textbf{0.600 $\pm$ 0.000}} & \cellcolor{lgray}{\textbf{0.600 $\pm$ 0.000}} & \cellcolor{lgray}{\textbf{0.533 $\pm$ 0.094}} & \cellcolor{lgray}{\textbf{0.644}} \\
\midrule
\midrule
\textbf{Model} & 2024.12 & 2025.01 & 2025.02 & 2025.03 & 2025.04 & 2025.05 & \textbf{Avg} \\
\midrule
XGBoost & 0.133 $\pm$ 0.094 & 0.000 $\pm$ 0.000 & 0.200 $\pm$ 0.000 & 0.000 $\pm$ 0.000 & 0.067 $\pm$ 0.094 & 0.000 $\pm$ 0.000 & 0.067 \\
SVR & 0.200 $\pm$ 0.000 & 0.133 $\pm$ 0.094 & 0.400 $\pm$ 0.000 & 0.400 $\pm$ 0.000 & 0.200 $\pm$ 0.000 & 0.200 $\pm$ 0.000 & 0.256 \\
Transformer & 0.333 $\pm$ 0.094 & 0.133 $\pm$ 0.094 & 0.333 $\pm$ 0.094 & 0.400 $\pm$ 0.000 & 0.400 $\pm$ 0.000 & 0.200 $\pm$ 0.000 & 0.300 \\
TGCN & 0.000 $\pm$ 0.000 & 0.200 $\pm$ 0.000 & 0.267 $\pm$ 0.094 & 0.200 $\pm$ 0.000 & 0.200 $\pm$ 0.000 & 0.133 $\pm$ 0.094 & 0.167 \\
\midrule
GPT-4.1-mini & 0.200 $\pm$ 0.000 & 0.000 $\pm$ 0.000 & 0.067 $\pm$ 0.094 & 0.000 $\pm$ 0.000 & 0.000 $\pm$ 0.000 & 0.000 $\pm$ 0.000 & 0.044 \\
GPT-4.1 & 0.267 $\pm$ 0.189 & 0.200 $\pm$ 0.000 & 0.200 $\pm$ 0.000 & 0.200 $\pm$ 0.000 & 0.133 $\pm$ 0.094 & 0.000 $\pm$ 0.000 & 0.167 \\
Gemini-3-Flash-Preview & - & 0.200 $\pm$ 0.000 & 0.267 $\pm$ 0.094 & 0.333 $\pm$ 0.094 & 0.333 $\pm$ 0.094 & 0.200 $\pm$ 0.000 & 0.267 \\
Gemini-3.1-Pro-Preview & - & 0.200 $\pm$ 0.000 & 0.400 $\pm$ 0.000 & 0.267 $\pm$ 0.094 & 0.200 $\pm$ 0.000 & 0.267 $\pm$ 0.094 & 0.267 \\
DeepSeek-V3.2 & - & 0.133 $\pm$ 0.094 & 0.267 $\pm$ 0.094 & 0.200 $\pm$ 0.000 & 0.200 $\pm$ 0.000 & 0.067 $\pm$ 0.094 & 0.173 \\
SciJudge-4B & - & 0.400 $\pm$ 0.000 & 0.200 $\pm$ 0.000 & 0.000 $\pm$ 0.000 & 0.067 $\pm$ 0.094 & 0.000 $\pm$ 0.000 & 0.136 \\
\midrule
\MethodName{} (Ours) & \cellcolor{lgray}{\textbf{0.600 $\pm$ 0.000}} & \cellcolor{lgray}{\textbf{0.667 $\pm$ 0.094}} & \cellcolor{lgray}{\textbf{0.800 $\pm$ 0.000}} & \cellcolor{lgray}{\textbf{0.600 $\pm$ 0.000}} & \cellcolor{lgray}{\textbf{0.667 $\pm$ 0.094}} & \cellcolor{lgray}{\textbf{0.533 $\pm$ 0.094}} & \cellcolor{lgray}{\textbf{0.644}} \\
\midrule
\textbf{Model} & 2025.06 & 2025.07 & 2025.08 & 2025.09 & 2025.10 & 2025.11 & \multicolumn{1}{c}{\textbf{Avg}} \\
\midrule
XGBoost & 0.000 $\pm$ 0.000 & 0.067 $\pm$ 0.094 & 0.000 $\pm$ 0.000 & 0.000 $\pm$ 0.000 & 0.000 $\pm$ 0.000 & 0.133 $\pm$ 0.094 & 0.033 \\
SVR & 0.200 $\pm$ 0.000 & 0.267 $\pm$ 0.094 & 0.200 $\pm$ 0.000 & 0.200 $\pm$ 0.000 & 0.200 $\pm$ 0.000 & 0.200 $\pm$ 0.000 & 0.211 \\
Transformer & 0.200 $\pm$ 0.000 & 0.267 $\pm$ 0.094 & 0.200 $\pm$ 0.163 & 0.400 $\pm$ 0.000 & 0.200 $\pm$ 0.000 & 0.200 $\pm$ 0.000 & 0.244 \\
TGCN & 0.067 $\pm$ 0.094 & 0.200 $\pm$ 0.000 & 0.200 $\pm$ 0.000 & 0.200 $\pm$ 0.000 & 0.200 $\pm$ 0.000 & 0.133 $\pm$ 0.094 & 0.167 \\
\midrule
GPT-4.1-mini & 0.400 $\pm$ 0.000 & 0.000 $\pm$ 0.000 & 0.133 $\pm$ 0.094 & 0.333 $\pm$ 0.094 & 0.400 $\pm$ 0.000 & 0.200 $\pm$ 0.000 & 0.244 \\
GPT-4.1 & 0.200 $\pm$ 0.000 & 0.067 $\pm$ 0.094 & 0.067 $\pm$ 0.094 & 0.000 $\pm$ 0.000 & 0.333 $\pm$ 0.094 & 0.133 $\pm$ 0.094 & 0.133 \\
Gemini-3-Flash-Preview & 0.200 $\pm$ 0.000 & 0.400 $\pm$ 0.000 & 0.400 $\pm$ 0.000 & 0.133 $\pm$ 0.094 & 0.200 $\pm$ 0.000 & 0.200 $\pm$ 0.000 & 0.256 \\
Gemini-3.1-Pro-Preview & 0.333 $\pm$ 0.249 & 0.467 $\pm$ 0.094 & 0.200 $\pm$ 0.000 & 0.133 $\pm$ 0.094 & 0.200 $\pm$ 0.163 & 0.133 $\pm$ 0.094 & 0.244 \\
DeepSeek-V3.2 & 0.267 $\pm$ 0.094 & 0.200 $\pm$ 0.000 & 0.200 $\pm$ 0.000 & 0.400 $\pm$ 0.000 & 0.333 $\pm$ 0.094 & 0.200 $\pm$ 0.000 & 0.267 \\
GPT-5.2 & - & - & 0.267 $\pm$ 0.094 & 0.067 $\pm$ 0.094 & 0.200 $\pm$ 0.163 & 0.267 $\pm$ 0.094 & 0.200 \\
SciJudge-4B & 0.200 $\pm$ 0.000 & 0.333 $\pm$ 0.094 & 0.133 $\pm$ 0.094 & 0.000 $\pm$ 0.000 & 0.400 $\pm$ 0.000 & 0.267 $\pm$ 0.094 & 0.222 \\
\midrule
\MethodName{} (Ours) & \cellcolor{lgray}{\textbf{0.800 $\pm$ 0.000}} & \cellcolor{lgray}{\textbf{0.733 $\pm$ 0.094}} & \cellcolor{lgray}{\textbf{0.600 $\pm$ 0.000}} & \cellcolor{lgray}{\textbf{0.600 $\pm$ 0.000}} & \cellcolor{lgray}{\textbf{0.467 $\pm$ 0.094}} & \cellcolor{lgray}{\textbf{0.333 $\pm$ 0.094}} & \cellcolor{lgray}{\textbf{0.589}} \\
\bottomrule
\bottomrule
\end{tabular}
 }
\end{table}

\clearpage
\section{Boosting LLM Evaluators with Manifold Integration}
\label{app:manifold_integration}

While standard Large Language Models struggle to independently evaluate scientific impact due to their static semantic context, their reasoning and critique capabilities can be significantly enhanced when grounded with the geometric signals produced by our \MethodName{} framework. 

\subsection{Integration Methodology}

To achieve this integration, we modify the standard LLM evaluation prompt to explicitly include the macroscopic evolutionary momentum calculated by our model. Specifically, we inject the fused manifold and naive embedding scores directly into the system prompt structure, allowing the LLM to utilize them as an auxiliary signal.

The baseline evaluation prompt (Prompt Template 2) is augmented by injecting the following \texttt{\{frontier\_section\}} into the LLM's context window prior to the instructions:

\begin{leftbar}
\begin{Verbatim}[
  breaklines=true,
  breaksymbolleft={},   % remove symbol at start of wrapped line
  breaksymbolright={}   % (optional) remove symbol at end too
]
manifold scores for this eval batch (same index order as the ideas list below): 
where score is the trained manifold prediction and z-scores are taken over this batch. Use as auxiliary signal, judge each idea on its own merits:
- index 0: {score_0}
- index 1: {score_1}
...
\end{Verbatim}
\end{leftbar}

\subsection{Performance Improvements}

By feeding the output of the manifold model back into the LLM as explicit context, the LLM is no longer forced to operate in a temporal vacuum. Instead, it utilizes the continuous-time spatiotemporal trajectory as an authoritative, macro-level reference point. This allows the LLM to effectively combine its inherent, microscopic textual critique capabilities (e.g., evaluating methodological soundness and clarity) with \MethodName{}'s macroscopic structural awareness. 
This hybrid evaluation paradigm effectively mitigates the centroid collapse vulnerability. As shown in Table \ref{tab:llm_add}, models augmented with \MethodName{} exhibit an average Spearman rank correlation increase of over $+0.150$ compared to their unaugmented counterparts. This synergistic integration confirms that while LLMs are currently poor standalone forecasters of scientific evolution, they function exceptionally well as hybrid evaluators when geometrically anchored to the forward momentum of a field.

\begin{table}[ht]
\centering
\caption{Performance improvements of baseline LLM evaluators when augmented with \MethodName{} manifold scores, improvements are shown in {\color{green!60!black}green}. The experiments are carried out for three times and the standard deviation of the performance is reported.}
\label{tab:llm_add}
\resizebox{1\linewidth}{!}{
\begin{tabular}{l|lll|l}
\toprule
\toprule 
\textbf{Model} & 2024.06 - 2024.12 & 2025.01 - 2025.07 & 2025.08 - 2025.11 & \textbf{Avg} \\
\midrule
GPT-4.1-mini & 0.213 $\pm$ 0.079 & 0.094 $\pm$ 0.055 & 0.177 $\pm$ 0.045 & 0.158 $\pm$ 0.083 \\
\cellcolor{llgray}\bgroup w/ \MethodName{}\egroup & \cellcolor{llgray}\bgroup 0.391 $\pm$ 0.099 \color{green!60!black}(+0.178 \(\uparrow\))\egroup & \cellcolor{llgray}\bgroup 0.251 $\pm$ 0.067 \color{green!60!black}(+0.157 \(\uparrow\))\egroup & \cellcolor{llgray}\bgroup 0.304 $\pm$ 0.155 \color{green!60!black}(+0.127 \(\uparrow\))\egroup & \cellcolor{llgray}\bgroup 0.317 $\pm$ 0.122 \color{green!60!black}(+0.159 \(\uparrow\))\egroup \\
\midrule
GPT-4.1 & 0.125 $\pm$ 0.092 & 0.025 $\pm$ 0.086 & 0.052 $\pm$ 0.147 & 0.070 $\pm$ 0.114 \\
\cellcolor{llgray}\bgroup w/ \MethodName{}\egroup & \cellcolor{llgray}\bgroup 0.394 $\pm$ 0.100 \color{green!60!black}(+0.269 \(\uparrow\))\egroup & \cellcolor{llgray}\bgroup 0.366 $\pm$ 0.092 \color{green!60!black}(+0.341 \(\uparrow\))\egroup & \cellcolor{llgray}\bgroup 0.290 $\pm$ 0.137 \color{green!60!black}(+0.238 \(\uparrow\))\egroup & \cellcolor{llgray}\bgroup 0.360 $\pm$ 0.113 \color{green!60!black}(+0.290 \(\uparrow\))\egroup \\
\midrule
Gemini-3-Flash-Preview & - & 0.198 $\pm$ 0.079 & 0.176 $\pm$ 0.068 & 0.190 $\pm$ 0.076 \\
\cellcolor{llgray}\bgroup w/ \MethodName{}\egroup & \cellcolor{llgray}\bgroup -\egroup & \cellcolor{llgray}\bgroup 0.399 $\pm$ 0.083 \color{green!60!black}(+0.201 \(\uparrow\))\egroup & \cellcolor{llgray}\bgroup 0.410 $\pm$ 0.105 \color{green!60!black}(+0.234 \(\uparrow\))\egroup & \cellcolor{llgray}\bgroup 0.403 $\pm$ 0.092 \color{green!60!black}(+0.213 \(\uparrow\))\egroup \\
\midrule
Gemini-3.1-Pro-Preview & - & 0.167 $\pm$ 0.111 & 0.173 $\pm$ 0.066 & 0.169 $\pm$ 0.097 \\
\cellcolor{llgray}\bgroup w/ \MethodName{}\egroup & \cellcolor{llgray}\bgroup -\egroup & \cellcolor{llgray}\bgroup 0.362 $\pm$ 0.105 \color{green!60!black}(+0.195 \(\uparrow\))\egroup & \cellcolor{llgray}\bgroup 0.382 $\pm$ 0.104 \color{green!60!black}(+0.209 \(\uparrow\))\egroup & \cellcolor{llgray}\bgroup 0.369 $\pm$ 0.105 \color{green!60!black}(+0.200 \(\uparrow\))\egroup \\
\midrule
DeepSeek-V3.2 & - & 0.221 $\pm$ 0.092 & 0.161 $\pm$ 0.044 & 0.199 $\pm$ 0.084 \\
\cellcolor{llgray}\bgroup w/ \MethodName{}\egroup & \cellcolor{llgray}\bgroup -\egroup & \cellcolor{llgray}\bgroup 0.367 $\pm$ 0.129 \color{green!60!black}(+0.146 \(\uparrow\))\egroup & \cellcolor{llgray}\bgroup 0.409 $\pm$ 0.124 \color{green!60!black}(+0.248 \(\uparrow\))\egroup & \cellcolor{llgray}\bgroup 0.382 $\pm$ 0.128 \color{green!60!black}(+0.183 \(\uparrow\))\egroup \\
\midrule
GPT-5.2 & - & - & 0.179 $\pm$ 0.061 & 0.179 $\pm$ 0.061 \\
\cellcolor{llgray}\bgroup w/ \MethodName{}\egroup & \cellcolor{llgray}\bgroup -\egroup & \cellcolor{llgray}\bgroup -\egroup & \cellcolor{llgray}\bgroup 0.394 $\pm$ 0.103 \color{green!60!black}(+0.215 \(\uparrow\))\egroup & \cellcolor{llgray}\bgroup 0.394 $\pm$ 0.103 \color{green!60!black}(+0.215 \(\uparrow\))\egroup \\
\bottomrule
\bottomrule
\end{tabular}}
\end{table}

\section{Ablation Study on Manifold Learning}
\label{app:ablation_on_loss}

To rigorously validate the contribution of each geometric objective within the \MethodName{} architecture, we systematically evaluated the model's predictive performance after independently removing the inspiration loss ($\mathcal{L}_{inspire}$), vanguard loss ($\mathcal{L}_{vanguard}$), and spine loss ($\mathcal{L}_{spine}$). Furthermore, we assessed the importance of our retrieve-and-verify pipeline by substituting the refined inspiration graph with either the full citation graph or a random inspiration graph. For the random inspiration graph variant, we explicitly dismantle true knowledge flow by randomly selecting 8 papers published earlier than the target paper to serve as its inspiration papers.

As detailed in Table \ref{tab:ablation}, the ablation of any individual geometric constraint from the \MethodName{} architecture results in a pronounced degradation in predictive performance across the entire 18-month sliding window evaluation. While the fully optimized \MethodName{} model consistently achieves average Spearman rank correlations exceeding 0.510 across all distinct temporal phases, removing the vanguard loss yields the most severe performance drop. This quantitative decline empirically confirms that localized impact normalization is a critical component of the framework; without explicitly encouraging vanguard papers to align with the forward momentum of the topic, the model fails to geometrically differentiate high-impact research from marginal publications, severely crippling downstream rank inference. Similarly, the omission of either the inspiration loss or the spine loss leads to substantial correlation drops. This underscores that both the micro-level modeling of directional knowledge flow and the macro-level temporal anchoring of topic trajectories are strictly necessary to maintain the structural integrity and predictive power of the spatiotemporal manifold. Furthermore, substituting the LLM-verified inspiration graph with the full citation graph results in a substantial performance decline. This empirically demonstrates that raw citation networks introduce excessive noise and that isolating genuine foundational knowledge flow is essential for accurate impact forecasting. Meanwhile, utilizing the random inspiration graph also results in a distinct degradation in performance, further validating the necessity of extracting genuine semantic inspiration over arbitrary or purely statistical historical linkages.
\begin{table}[ht]
\centering
\caption{Quantitative ablation study evaluating the impact of each geometric objective. Removing the inspiration loss, vanguard loss, or spine loss results in pronounced degradation in Spearman rank correlation between predicted score and true impact weight compared to the full \MethodName{} model. The experiments are carried out for three times and the standard deviation of the performance is reported.}
\label{tab:ablation}
\resizebox{1\linewidth}{!}{
\begin{tabular}{l|cccccc|c}
\toprule
\toprule 
\textbf{Model} & 2024.06 & 2024.07 & 2024.08 & 2024.09 & 2024.10 & 2024.11 & \textbf{Avg} \\
\midrule
MLP (raw text embedding) & 0.058 $\pm$ 0.110 & 0.204 $\pm$ 0.050 & 0.238 $\pm$ 0.018 & 0.278 $\pm$ 0.076 & 0.255 $\pm$ 0.040 & 0.125 $\pm$ 0.019 & 0.193 \\
\midrule
w/ full citation graph & 0.160 $\pm$ 0.040 & 0.011 $\pm$ 0.109 & 0.193 $\pm$ 0.020 & 0.316 $\pm$ 0.058 & 0.167 $\pm$ 0.008 & 0.345 $\pm$ 0.050 & 0.199 \\
w/ random inspiration graph & 0.077 $\pm$ 0.091 & 0.099 $\pm$ 0.051 & 0.057 $\pm$ 0.121 & 0.258 $\pm$ 0.055 & 0.157 $\pm$ 0.040 & 0.128 $\pm$ 0.034 & 0.129 \\
\midrule
w/o inspiration loss & 0.160 $\pm$ 0.026 & 0.177 $\pm$ 0.045 & 0.185 $\pm$ 0.045 & 0.330 $\pm$ 0.037 & 0.228 $\pm$ 0.036 & 0.177 $\pm$ 0.039 & 0.209 \\
w/o vanguard loss & 0.061 $\pm$ 0.103 & 0.073 $\pm$ 0.061 & 0.223 $\pm$ 0.062 & 0.235 $\pm$ 0.047 & 0.235 $\pm$ 0.054 & 0.108 $\pm$ 0.061 & 0.156 \\
w/o spine loss & 0.186 $\pm$ 0.036 & 0.210 $\pm$ 0.055 & 0.276 $\pm$ 0.016 & 0.396 $\pm$ 0.036 & 0.307 $\pm$ 0.027 & 0.234 $\pm$ 0.021 & 0.268 \\
\midrule
\MethodName{} & \cellcolor{lgray}{\textbf{0.521 $\pm$ 0.051}} & \cellcolor{lgray}{\textbf{0.633 $\pm$ 0.016}} & \cellcolor{lgray}{\textbf{0.483 $\pm$ 0.006}} & \cellcolor{lgray}{\textbf{0.531 $\pm$ 0.007}} & \cellcolor{lgray}{\textbf{0.562 $\pm$ 0.026}} & \cellcolor{lgray}{\textbf{0.482 $\pm$ 0.036}} & \cellcolor{lgray}{\textbf{0.535}} \\
\midrule
\midrule
\textbf{Model} & 2024.12 & 2025.01 & 2025.02 & 2025.03 & 2025.04 & 2025.05 & \textbf{Avg} \\
\midrule
MLP (raw text embedding) & 0.324 $\pm$ 0.035 & 0.281 $\pm$ 0.050 & 0.457 $\pm$ 0.022 & 0.274 $\pm$ 0.083 & 0.250 $\pm$ 0.052 & 0.437 $\pm$ 0.051 & 0.337 \\
\midrule
w/ full citation graph & 0.340 $\pm$ 0.028 & 0.014 $\pm$ 0.027 & 0.263 $\pm$ 0.003 & 0.360 $\pm$ 0.002 & 0.328 $\pm$ 0.026 & 0.474 $\pm$ 0.008 & 0.296 \\
w/ random inspiration graph & 0.167 $\pm$ 0.047 & 0.214 $\pm$ 0.020 & 0.383 $\pm$ 0.006 & 0.062 $\pm$ 0.083 & 0.279 $\pm$ 0.027 & 0.440 $\pm$ 0.034 & 0.257 \\
\midrule
w/o inspiration loss & 0.295 $\pm$ 0.080 & 0.261 $\pm$ 0.034 & 0.421 $\pm$ 0.031 & 0.216 $\pm$ 0.121 & 0.332 $\pm$ 0.039 & 0.485 $\pm$ 0.024 & 0.335 \\
w/o vanguard loss & 0.256 $\pm$ 0.037 & 0.115 $\pm$ 0.011 & 0.266 $\pm$ 0.036 & 0.145 $\pm$ 0.091 & 0.179 $\pm$ 0.037 & 0.224 $\pm$ 0.086 & 0.197 \\
w/o spine loss & 0.346 $\pm$ 0.035 & 0.253 $\pm$ 0.042 & 0.436 $\pm$ 0.014 & 0.261 $\pm$ 0.052 & 0.372 $\pm$ 0.009 & 0.529 $\pm$ 0.024 & 0.366 \\
\midrule
\MethodName{} & \cellcolor{lgray}{\textbf{0.524 $\pm$ 0.010}} & \cellcolor{lgray}{\textbf{0.433 $\pm$ 0.055}} & \cellcolor{lgray}{\textbf{0.563 $\pm$ 0.025}} & \cellcolor{lgray}{\textbf{0.492 $\pm$ 0.023}} & \cellcolor{lgray}{\textbf{0.450 $\pm$ 0.005}} & \cellcolor{lgray}{\textbf{0.617 $\pm$ 0.029}} & \cellcolor{lgray}{\textbf{0.513}} \\
\midrule
\midrule
\textbf{Model} & 2025.06 & 2025.07 & 2025.08 & 2025.09 & 2025.10 & 2025.11 & \textbf{Avg} \\
\midrule
MLP (raw text embedding) & 0.361 $\pm$ 0.018 & 0.248 $\pm$ 0.056 & 0.363 $\pm$ 0.095 & 0.445 $\pm$ 0.026 & 0.204 $\pm$ 0.064 & -0.062 $\pm$ 0.082 & 0.260 \\
\midrule
w/ full citation graph & 0.345 $\pm$ 0.042 & 0.313 $\pm$ 0.032 & 0.356 $\pm$ 0.015 & 0.361 $\pm$ 0.027 & 0.224 $\pm$ 0.098 & 0.235 $\pm$ 0.078 & 0.306 \\
w/ random inspiration graph & 0.364 $\pm$ 0.009 & 0.194 $\pm$ 0.025 & 0.275 $\pm$ 0.067 & 0.317 $\pm$ 0.012 & 0.131 $\pm$ 0.104 & 0.107 $\pm$ 0.050 & 0.231 \\
\midrule
w/o inspiration loss & 0.389 $\pm$ 0.021 & 0.292 $\pm$ 0.085 & 0.357 $\pm$ 0.050 & 0.367 $\pm$ 0.036 & 0.295 $\pm$ 0.124 & 0.206 $\pm$ 0.053 & 0.317 \\
w/o vanguard loss & 0.260 $\pm$ 0.034 & 0.213 $\pm$ 0.034 & 0.315 $\pm$ 0.029 & 0.414 $\pm$ 0.036 & 0.343 $\pm$ 0.123 & 0.156 $\pm$ 0.134 & 0.283 \\
w/o spine loss & 0.463 $\pm$ 0.042 & 0.367 $\pm$ 0.026 & 0.454 $\pm$ 0.012 & 0.436 $\pm$ 0.014 & 0.365 $\pm$ 0.047 & 0.269 $\pm$ 0.075 & 0.392 \\
\midrule
\MethodName{} & \cellcolor{lgray}{\textbf{0.616 $\pm$ 0.026}} & \cellcolor{lgray}{\textbf{0.676 $\pm$ 0.027}} & \cellcolor{lgray}{\textbf{0.582 $\pm$ 0.013}} & \cellcolor{lgray}{\textbf{0.552 $\pm$ 0.032}} & \cellcolor{lgray}{\textbf{0.399 $\pm$ 0.023}} & \cellcolor{lgray}{\textbf{0.370 $\pm$ 0.052}} & \cellcolor{lgray}{\textbf{0.532}} \\
\bottomrule
\bottomrule
\end{tabular}}
\end{table}

\clearpage
\section{Sensitivity Analysis of Loss Weights}
\label{app:loss_w_sensitivity}
To rigorously evaluate the stability of the \MethodName{} framework and the interplay between its constituent geometric constraints, we conduct a comprehensive sensitivity analysis of the loss formulation hyperparameters. Specifically, we systematically vary the relative weights of the spine binding loss ($\alpha$), inspiration loss ($\beta$), and vanguard loss ($\gamma$) within the total objective function to explore the combinatorial hyperparameter space. The resultant predictive performance, quantified via the Spearman rank correlation between the predicted and ground-truth impact weights, is mapped across these varying weight distributions.  Specifically, the heatmap indicates that the framework consistently achieves strong predictive performance when the inspiration loss weight ($\beta$) is kept relatively small, provided that the spine binding loss ($\alpha$) and vanguard loss ($\gamma$) are located within a reasonable range.
This analysis elucidates the optimal balance required among the three structural objectives, further corroborating the findings from our ablation studies that the concurrent application of macroscopic temporal anchoring, directional knowledge flow modeling, and localized impact normalization is imperative for maximizing the model's forecasting efficacy.
\begin{figure}[ht]
    \centering
    \includegraphics[width=\linewidth]{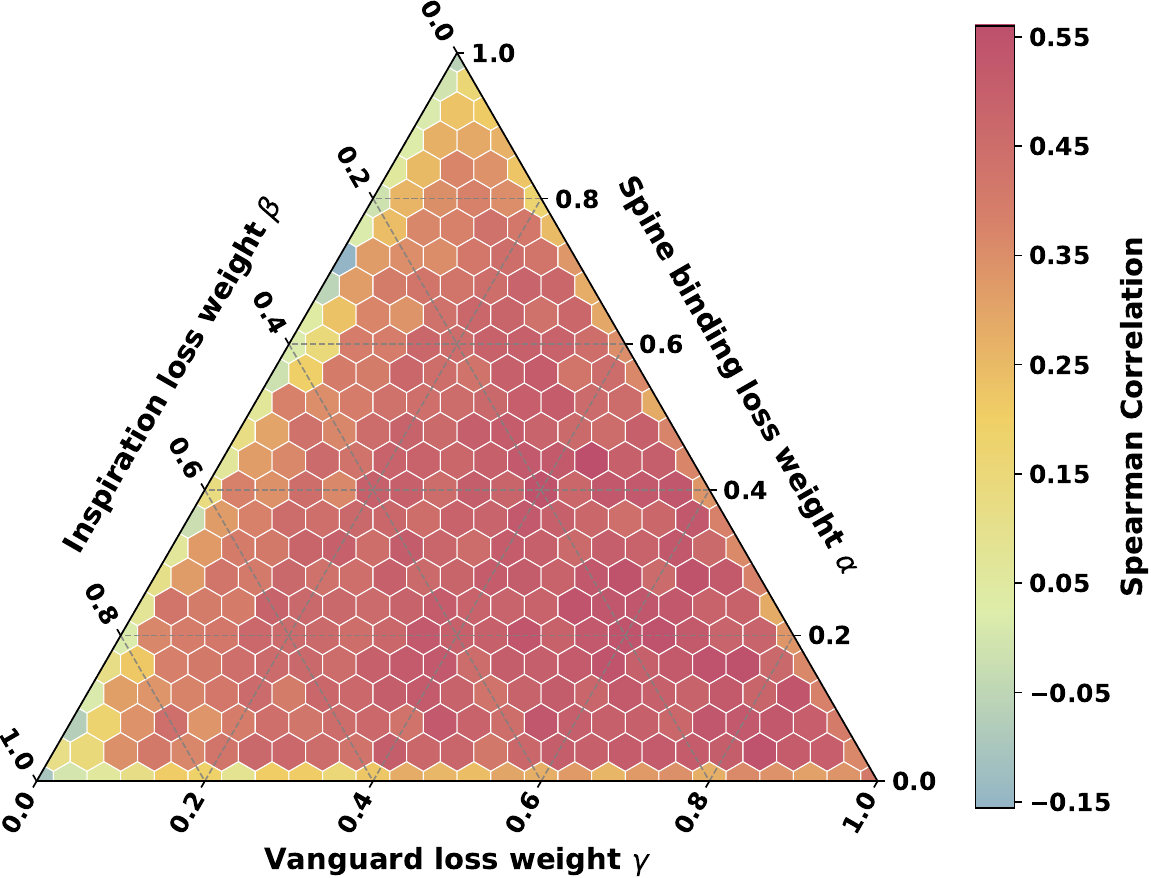}
    \caption{Ternary heatmap illustrating the sensitivity of the \MethodName{} framework's predictive performance to variations in the geometric loss weights. The color intensity corresponds to the Spearman rank correlation achieved across different combinatorial assignments of the spine binding loss weight ($\alpha$), inspiration loss weight ($\beta$), and vanguard loss weight ($\gamma$). The visualization indicates the optimal operating region where all three structural constraints are synergistically balanced to structure the spatiotemporal manifold.}
    \label{fig:loss_w_sensitivity}
\end{figure}

\clearpage
\section{More Case On Text Embeddings vs. Manifold Embeddings}
\label{app:text_vs_manifold}

We provide more cases on comparing using the original text embeddings of the papers produced by the text embedding model and the manifold embeddings produced by \MethodName{}.

\begin{figure}[ht]
    \centering
    \includegraphics[width=\linewidth]{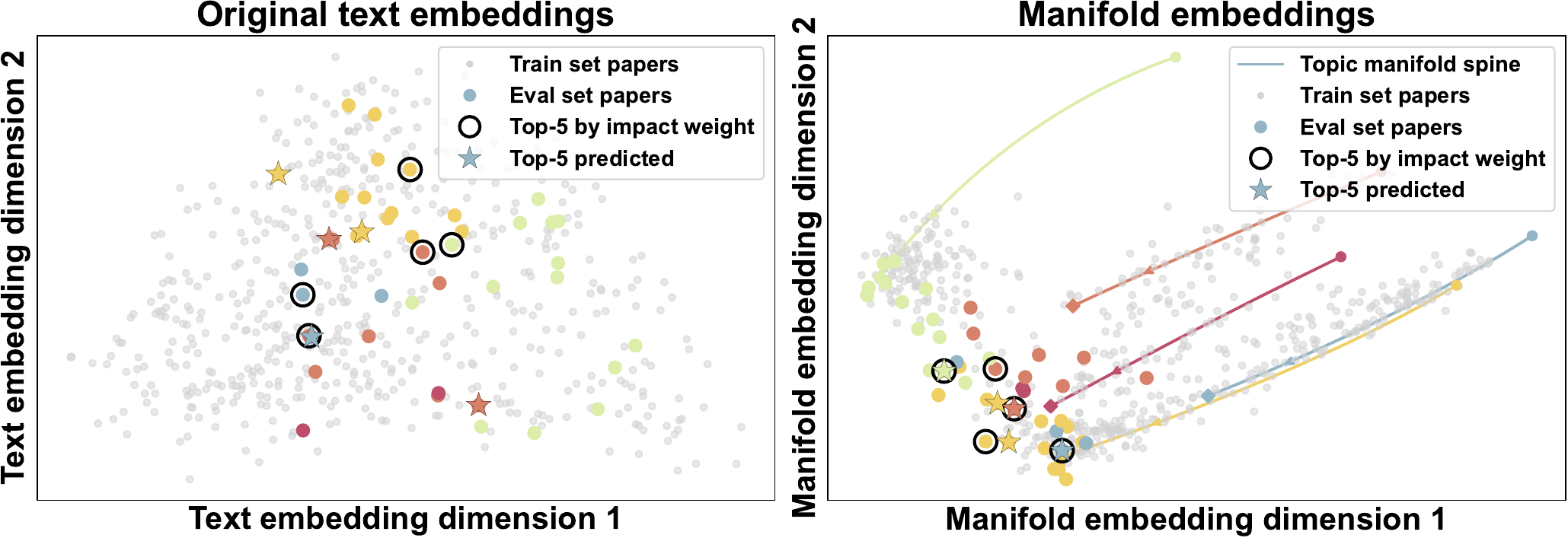}
    \caption{Additional examples comparing original text embedding distributions against \MethodName{} manifold embeddings.}
    \label{fig:raw_vs_manifold_1}
\end{figure}

\begin{figure}[ht]
    \centering
    \includegraphics[width=\linewidth]{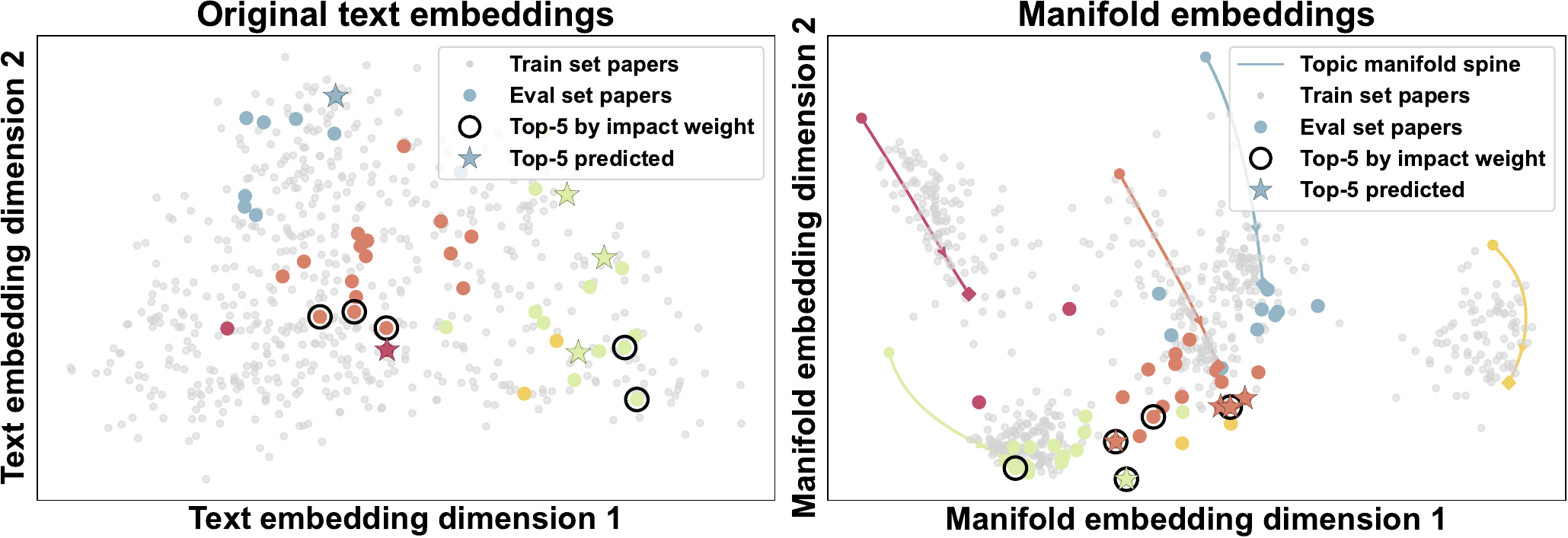}
    \caption{Additional examples comparing original text embedding distributions against \MethodName{} manifold embeddings.}
    \label{fig:raw_vs_manifold_2}
\end{figure}

\begin{figure}[ht]
    \centering
    \includegraphics[width=\linewidth]{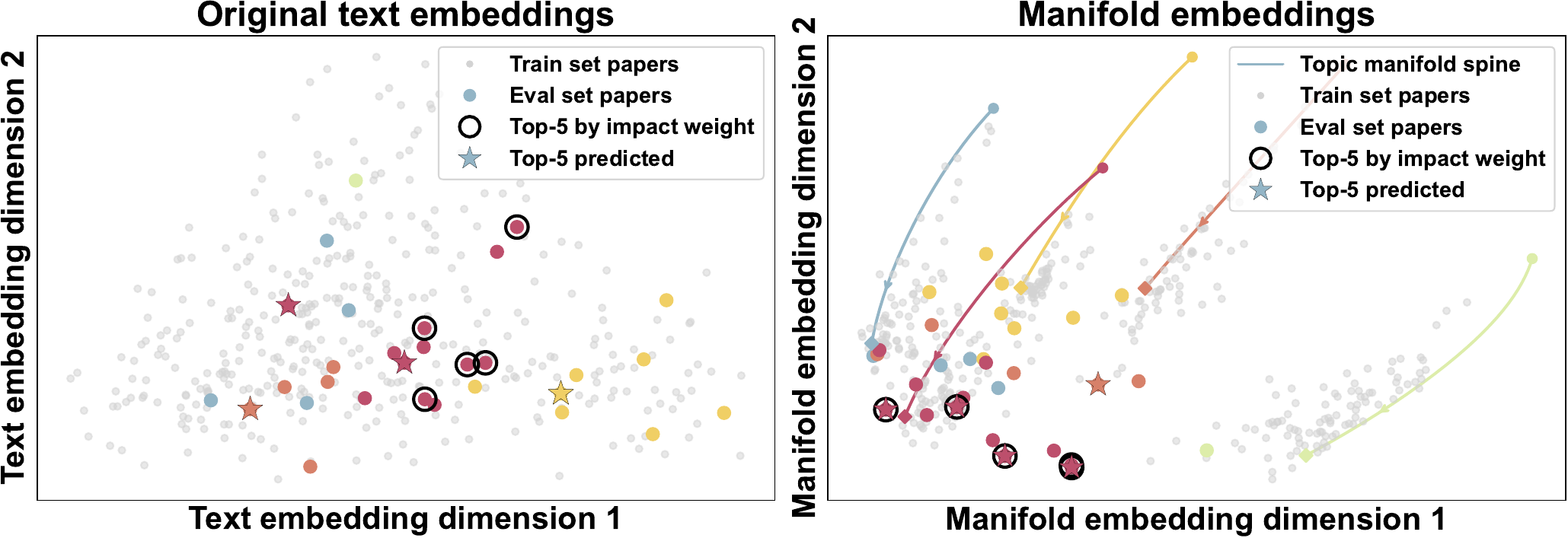}
    \caption{Additional examples comparing original text embedding distributions against \MethodName{} manifold embeddings.}
    \label{fig:raw_vs_manifold_3}
\end{figure}

\newpage
\section{Correlation of Normalized Impact Score Components with Human Ratings}
\label{app:corr_human}

Detailed scatter plots demonstrate a strong alignment between our impact score and human ratings across three research domains of 300 papers. Plotting human ratings (x-axis) against metric scores (y-axis) alongside dashed linear regression lines and Pearson correlations ($\rho$), reveals that the composite impact score captures human consensus significantly better than any individual normalized component.
\begin{figure}[ht]
    \centering
    \includegraphics[width=\linewidth]{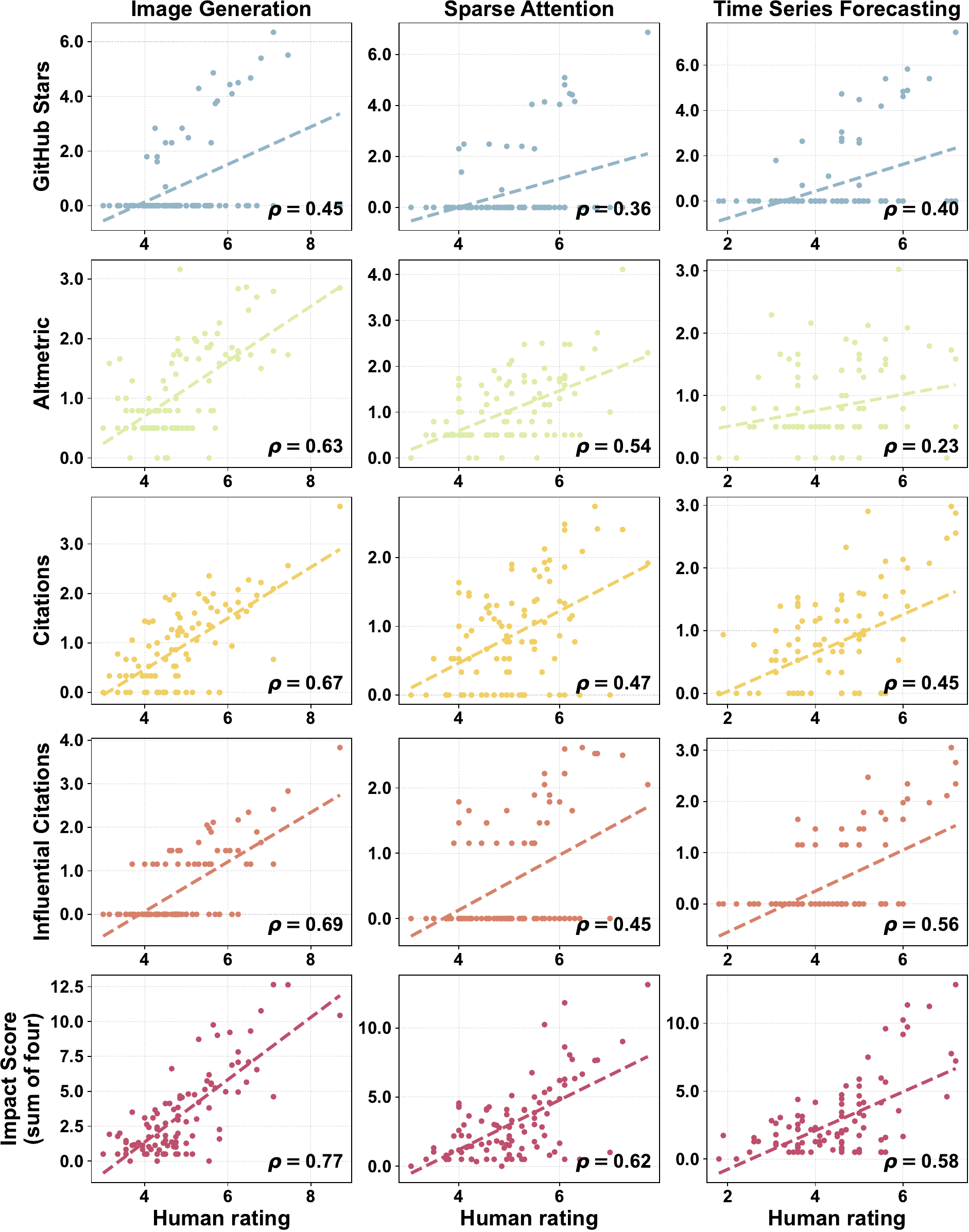}
    \caption{Correlation of normalized impact score components with human ratings. The composite impact score correlates more strongly with human consensus than any individual component.}
    \label{fig:corr_human}
\end{figure}

% \clearpage
% \input{checklist.tex}

\end{document}